\documentclass[11pt]{article}

\PassOptionsToPackage{hyperfootnotes=false}{hyperref}
\usepackage[preprint]{acl}
\usepackage{times}
\usepackage{latexsym}
\usepackage[T1]{fontenc}
\usepackage[utf8]{inputenc}
\usepackage{microtype}
\usepackage{inconsolata}
\usepackage{graphicx}
\usepackage{caption}
\usepackage{placeins}
\usepackage{colortbl}
\usepackage{tabularx}
\definecolor{acedrow}{RGB}{239,245,248}

\usepackage{booktabs}
\usepackage{multirow}
\usepackage{amsmath}
\usepackage{amssymb}
\usepackage{array}
\usepackage{float}
\floatstyle{ruled}
\newfloat{algorithm}{tbp}{loa}
\floatname{algorithm}{Algorithm}
\newcolumntype{L}[1]{>{\raggedright\arraybackslash}p{#1}}

\newcommand{\method}{ACED}
\newcommand{\cand}{ACED-CANDIDATE}
\newcommand{\allaced}{ACED-ALL}
\newcommand{\psc}{PoT}
\newcommand{\scbase}{SC}
\newcommand{\prove}{ProVe}
\newcommand{\acr}{ACR}

\newcommand{\anl}{\mathcal{A}_{\mathrm{NL}}}
\newcommand{\ap}{\mathcal{A}_{P}}
\newcommand{\ocand}{\Omega_{\mathrm{cand}}}
\newcommand{\oall}{\Omega_{\mathrm{all}}}

\title{ACED: Reference-Anchored Aggregation of Natural-Language and Executable Evidence for Mathematical Reasoning}

\author{
  Yiyi Lu\textsuperscript{1}\thanks{Equal contribution.}\quad
  Jiaxiang Geng\textsuperscript{1,2}\footnotemark[1]\quad
  Bing Luo\textsuperscript{1}\thanks{Corresponding author.}\\
  \textsuperscript{1}Duke Kunshan University\quad
  \textsuperscript{2}The University of Hong Kong\\
  \texttt{\{yl996,jg645,bl291\}@duke.edu}
}

\hypersetup{
  pdftitle={ACED: Reference-Anchored Aggregation of Natural-Language and Executable Evidence for Mathematical Reasoning},
  pdfauthor={Yiyi Lu, Jiaxiang Geng, Bing Luo}
}

\begin{document}
\maketitle

\begin{abstract}
Natural-language self-consistency and executable programs provide complementary test-time evidence for mathematical reasoning, yet either channel may support the same semantic error. Thus we propose \textbf{ACED}, a reference-anchored framework that treats the self-consistency answer as the default and executable outputs as supporting evidence rather than interchangeable votes. ACED converts source-specific answer counts into smoothed supports, combines them through a weighted logarithmic pool, and overrides the reference only when a calibrated relative-evidence margin is exceeded. The same rule supports both candidate selection and admission of executable-only answers. Across five math word-problem benchmarks, ACED matches or outperforms the corresponding baselines. On GSM8K, ACED-CANDIDATE recovers 89.7\% of the attainable candidate-set gain with only three damaging overrides, while ACED-ALL reaches 86.9\% accuracy versus 82.6\% for direct Program-of-Thought aggregation. These results show that preserving evidence provenance enables more reliable integration of natural-language and executable reasoning.
\end{abstract}

\begin{figure*}[t]
\centering
\includegraphics[width=0.92\textwidth]{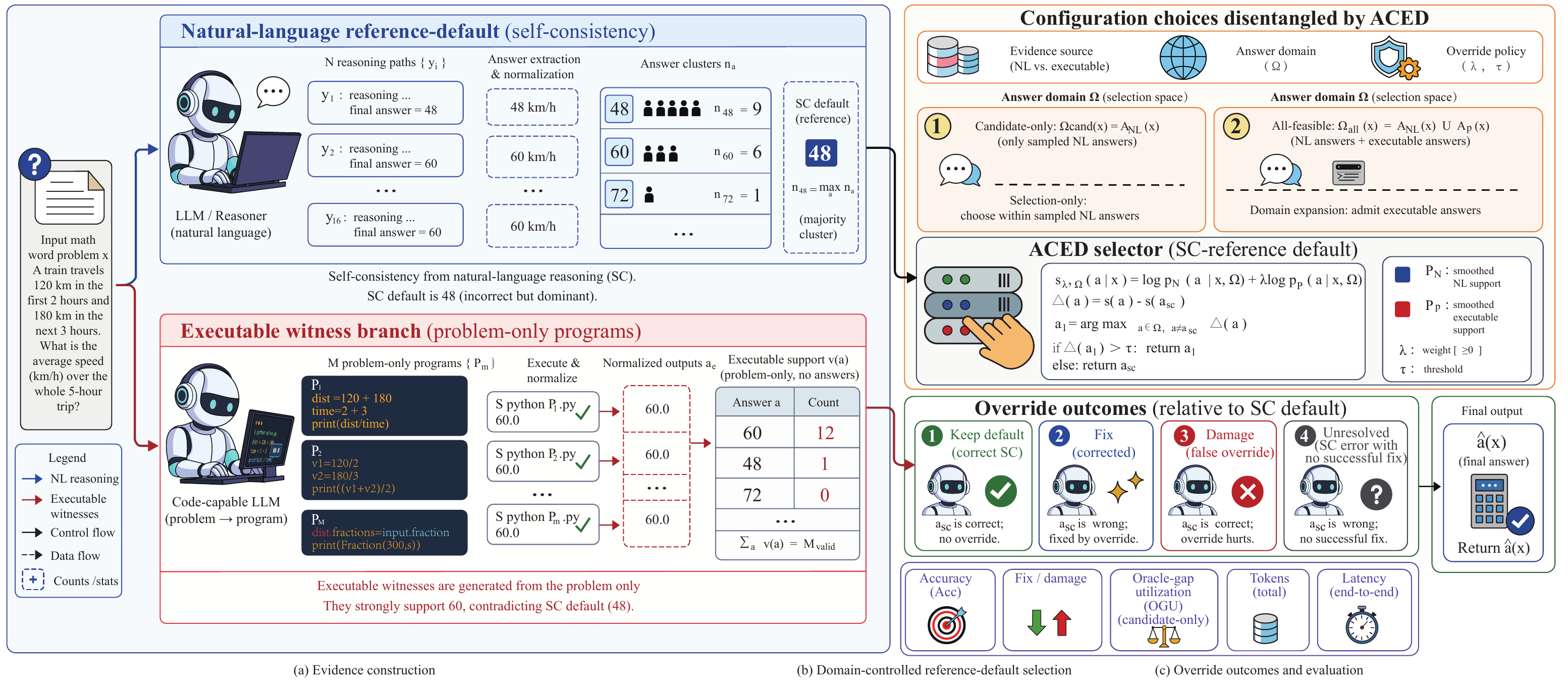}
\caption{Overview of \method{}. 
(a) Self-consistency offers the reference answer and problem-only programs provide a separate evidence source.
(b) The selector operates over candidate-only and all-feasible answer domains.
(c) Outcomes distinguish retained correct references, successful fixes, damaging replacements, and unresolved errors.}
\label{fig:overview}
\end{figure*}

\section{Introduction}
Large Language Models (LLMs) have achieved substantial progress in mathematical reasoning, driven not only by improved model capabilities but also by increased computation at inference time. Rather than committing to a single directly generated response, modern reasoning systems increasingly adopt a generate-and-select paradigm: they produce multiple candidate reasoning trajectories and aggregate or verify them through answer consistency, learned verifiers, or process-level feedback \citep{wei2022cot,wang2023selfconsistency,cobbe2021gsm8k,lightman2024verify}. This paradigm has been further advanced by test-time scaling methods that allocate additional inference computation to sampling, search, and verification \citep{brown2024largelanguagemonkeys,snell2024testtimecompute}. Consequently, the reliability of mathematical reasoning depends not only on whether a correct solution can be generated, but also on whether it can be reliably identified from a collection of diverse and potentially conflicting test-time evidence.
 
Two complementary sources of test-time evidence have been widely explored for mathematical reasoning. \textit{Self-consistency methods}, a natural-language reasoning way, represent a solution as an explicit sequence of intermediate reasoning steps and estimate answer reliability from the agreement among multiple sampled trajectories \citep{wei2022cot,wang2023selfconsistency}. In contrast, \textit{executable reasoning methods} prompt LLMs to generate programs or interact with external tools, delegating symbolic and numerical computation to interpreters or specialized tool environments \citep{gao2023pal,chen2023programofthoughts,gou2024tora}. These two forms of evidence provide complementary signals: natural-language trajectories expose the model's interpretation and decomposition of the problem, whereas executable outputs provide reproducible results. Their combination thus offers a promising means of improving answer selection beyond relying on either reasoning channel alone.

However, neither agreement among natural-language trajectories nor successful program execution guarantees semantic correctness. Self-consistency treats repeated answers as stronger evidence, yet trajectories sampled from the same model may share the same misunderstanding of a problem, causing additional sampling to reinforce an incorrect majority rather than recover a correct minority answer \citep{brown2024largelanguagemonkeys,mirzadeh2025gsmsymbolic}. Executable reasoning can reduce arithmetic and symbolic errors by offloading computation to an interpreter. This advantage, however, presupposes that code generation has correctly translated the original problem. When the model misinterprets the task, it may produce an executable but semantically incorrect solution, and the interpreter will faithfully return the answer to that mistaken formulation.

These observations make the integration of natural-language and executable evidence fundamentally nontrivial. The two channels provide complementary signals, yet either may support a semantically incorrect answer. An effective aggregation method must therefore exploit their complementary strengths while accounting for their distinct failure modes. This calls for a principled decision rule that reconciles potentially conflicting evidence from both channels to determine the final answer.

To this end, we propose \method{}, which formulates the evidence integration as a reference-relative override problem. The self-consistency prediction serves as the reference, while a code-capable model independently generates executable witnesses without observing the reference or the sampled natural-language trajectories. \method{} converts answer frequencies from both channels into smoothed supports, fuses them through a weighted logarithmic pool, and evaluates each alternative by its log support-ratio margin over the reference. Both the executable-evidence weight and the override threshold are selected on held-out calibration data and frozen before test evaluation. An alternative replaces the reference only when its margin exceeds this calibrated threshold, allowing a sufficiently supported alternative to correct an erroneous self-consistency majority while retaining the reference when replacement is not justified.

Our main contributions can be summarized as follows:
\begin{itemize}
    \setlength{\itemsep}{0.1em}
    \item We formalize executable aggregation as a reference-relative override problem, explicitly positioning self-consistency-derived answers as incumbents and executable outputs as supporting evidence rather than direct voters.
    \item We introduce \method{}, a transparent and interpretable calibrated support-ratio selector that systematically evaluates the sufficiency of executable evidence required to override the default incumbent answer.
    \item We clearly distinguish between answer selection and answer admission by evaluating \method{} in two distinct contexts, namely \cand{} and \allaced{}, thus isolating whether executable evidence primarily improves candidate selection among existing solutions or facilitates the safe introduction of novel executable-only answers.
    \item We propose a comprehensive, damage-aware evaluation framework, incorporating metrics for beneficial overrides (fixes), detrimental overrides (false overrides), ablation analyses, robustness tests against stronger incumbents, transferability evaluations, and cost assessments. The findings underscore that executable evidence significantly enhances reasoning accuracy when employed as calibrated supporting evidence rather than as default decision-makers.
\end{itemize}

\section{Related Work}

\paragraph{Selection and calibration of sampled reasoning.} Chain-of-thought prompting and self-consistency improve reasoning by generating multiple solution trajectories and aggregating their final answers \citep{wei2022cot,wang2023selfconsistency}. Later work strengthens this generate-and-select paradigm through confidence calibration, learned verifiers, and adaptive allocation of test-time computation \citep{cobbe2021gsm8k,lightman2024verify, wang2024selfconsistencycalibration,jeong2025acr, aggarwal-etal-2023-lets,snell2024testtimecompute}. These approaches mainly address which sampled solution should be trusted or whether additional evidence should be generated, typically through candidate ranking, filtering, or confidence estimation. \method{} takes a different view of the final decision: the self-consistency prediction serves as the reference and is replaced only when independent evidence provides sufficient support for an alternative. This formulation makes successful corrections and harmful overrides explicit within the same decision rule.

\paragraph{Executable and heterogeneous reasoning evidence.} Program-aided methods such as PAL and Program of Thoughts offload computation to executable code, while later work studies when programmatic reasoning is effective and how its confidence differs from natural-language reasoning \citep{gao2023pal,chen2023programofthoughts,bi2023programthoughts,kabra2024program}. Other approaches combine natural-language and executable reasoning through dynamic modality selection, candidate-conditioned program verification, or multi-source ensembling and tool aggregation \citep{toh2024prove,xiong2024incmath,yao2025multitag,saparkhan2026cotpot}. These studies establish the complementary value of natural-language and executable reasoning, but primarily use programs to generate, verify, or symmetrically ensemble solutions. \method{} focuses instead on how conflicting evidence should determine the final answer: independently generated executable witnesses provide source-specific support for retaining the self-consistency reference, selecting another sampled answer, or admitting an executable-only answer.

\section{Preliminaries and Problem Setup}

\paragraph{Motivating conflict.}
Table~\ref{tab:motivating-conflict} illustrates a simple conflict between the two evidence sources. With ten samples from each channel, source-agnostic pooling favors \(A\) by 9 to 8, although \(B\) is supported by both channels. With add-one smoothing and equal executable weight \((\lambda=1)\), the product support instead favors \(B\), 16 to 10, with margin \(\log(16/10)\). This reversal does not imply that \(B\) is correct, while it motivates calibrating both the executable weight and the margin required to replace the reference.

\begin{table}[htbp]
\centering
\small
\setlength{\tabcolsep}{5.5pt}
\begin{tabular}{@{}lccc@{}}
\toprule
Answer & \shortstack{NL\\support} & \shortstack{Program\\support} &
\shortstack{Pooled\\count} \\
\midrule
\(A\) & 9 & 0 & 9 \\
\(B\) & 1 & 7 & 8 \\
\(C,D,E\) (each) & 0 & 1 & 1 \\
\bottomrule
\end{tabular}
\caption{A minimal conflict in which pooled voting favors \(A\), while
add-one product support favors \(B\).}
\label{tab:motivating-conflict}
\end{table}
\FloatBarrier

\paragraph{Formal setup.}
Let \(x\) be a math problem and \(a^\star\) its ground-truth answer, used only for calibration and evaluation. From \(N\) normalized natural-language answers, let \(n_a\) denote the frequency of \(a\) and \(\anl(x)=\{a:n_a>0\}\). The self-consistency reference is \begin{equation}
a_0(x)=a_{\mathrm{SC}}(x)
=
\arg\max_{a\in\anl(x)} n_a,
\end{equation}
with deterministic tie-breaking. It serves as the reference but is not
assumed correct.

Independently, a code model generates \(M\) problem-only Python witnesses from \(x\), without access to candidate answers, \(a_0\), natural-language traces, or labels. After label-free validation and normalization, \(M_{\mathrm{valid}}\) denotes the number of retained executions, \(v(a)\) the number returning \(a\), and \(\ap(x)=\{a:v(a)>0\}\) the executable answer set. The appendices provide the complete output-accounting convention.

\paragraph{Answer domains.}
We consider two admissible answer domains:
\begin{equation}
\begin{aligned}
\ocand(x)&=\anl(x),\\
\oall(x)&=\anl(x)\cup\ap(x).
\end{aligned}
\end{equation}
\cand{} restricts selection to sampled natural-language answers, whereas
\allaced{} additionally admits executable-only answers. Both use the same evidence-pooling and acceptance rule.

\section{Method}
\label{sec:aced-method}

\method{} treats natural-language reasoning and executable witnessing as complementary but imperfect evidence channels. Natural-language trajectories interpret the problem and its requested quantity but may make arithmetic, counting, or algebraic errors. Programs can execute a computation correctly while consistently targeting the wrong quantity, unit, denominator, or scope. We therefore retain each source's support rather than treating their outputs as interchangeable votes, and ask whether the combined evidence justifies replacing the self-consistency reference. We first express this comparison through source-level supports and an information quotient, then construct the fused support needed to turn it into a calibrated decision.

\paragraph{Source-level supports.}
For an admissible domain $\Omega(x)$, we set unobserved answer counts to zero and define the smoothed supports
\begin{align}
p_N(a\mid x,\Omega)
&=\frac{n_a+\beta_N}{N+\beta_N|\Omega(x)|},
\label{eq:aced-support-n}\\
p_P(a\mid x,\Omega)
&=\frac{v(a)+\beta_P}
        {M_{\mathrm{valid}}+\beta_P|\Omega(x)|}.
\label{eq:aced-support-p}
\end{align}
Here $\beta_N,\beta_P>0$, with $\beta_N=\beta_P=1$ in the main experiments. These supports summarize observed frequencies rather than calibrated probabilities of correctness and require no independence assumption on the samples.

For the KL construction, we use normalized source distributions $p_N^+,p_P^+$ on an augmented answer space $S_\Omega$. Retained nonnumeric outputs enter a non-selectable null bucket $\varnothing$, and under $\ocand$, executable answers outside the candidate set enter an additional bucket $\bot$. Both buckets have zero natural-language counts. The resulting distributions preserve all support ratios between admissible answers within each source (Appendix~\ref{sec:augmented-supports}).

\paragraph{Relative evidence and the information quotient.}
The supports above describe each source separately, whereas a replacement decision requires comparing answers using both. For a positive fused support $q(a\mid x,\Omega)$, we define the \emph{information quotient} of answer $a$ relative to $b$ as
\begin{equation}
\mathcal Q_\Omega(a;b\mid x)
=\frac{q(a\mid x,\Omega)}{q(b\mid x,\Omega)}.
\label{eq:aced-information-quotient}
\end{equation}
The logarithm of this quotient gives an additive evidence margin. The quotient measures relative support, not a calibrated likelihood ratio or Bayes factor. Computing it requires a fused support $q$, which we construct from the two source distributions below.

\paragraph{KL least-distortion evidence pooling.}
A linear pool combines source support additively, allowing strong support from one channel to offset weak support from the other. We instead construct the fused support by minimizing its weighted KL distortion from the two source distributions:
\begin{equation}
q_\lambda
=\arg\min_{q\in\Delta(S_\Omega)}
\mathrm{KL}(q\Vert p_N^+)
+\lambda\,\mathrm{KL}(q\Vert p_P^+),
\label{eq:aced-kl-pool}
\end{equation}
where $\lambda\ge0$ controls the contribution of executable evidence. The direction $\mathrm{KL}(q\Vert p)$ penalizes assigning substantial fused mass to answers with weak source support, while smoothing keeps the divergences finite. The solution is a logarithmic pool. For admissible answers, source normalizers contribute only answer-independent factors, giving
\begin{equation}
\begin{aligned}
q_\lambda(a)
&\propto p_N(a\mid x,\Omega)^{\frac{1}{1+\lambda}}
          p_P(a\mid x,\Omega)^{\frac{\lambda}{1+\lambda}},\\
\widetilde q_\lambda(a)
&=p_N(a\mid x,\Omega)\,p_P(a\mid x,\Omega)^\lambda.
\end{aligned}
\label{eq:aced-logarithmic-pool}
\end{equation}
The second line defines the unscaled support used for selection. For fixed $\lambda$, its log-ratios are $1+\lambda$ times those of $q_\lambda$. Calibrating the threshold on this scale therefore preserves the same family of threshold decisions. The closed-form solution and scale conversion are derived in Appendices~\ref{sec:kl-derivation} and~\ref{sec:potential-interpretation}.

\paragraph{Evidence potential.}
If the correctness posterior were available, we could compare answers through their posterior support. Here, however, we have only two sets of observations that may be correlated and share systematic errors. We therefore use the KL-pooled support as a computable surrogate and define its logarithm as the \emph{evidence potential}:
\begin{equation}
\begin{aligned}
s_{\lambda,\Omega}(a;x)
&=\log\widetilde q_\lambda(a)\\
&=\log p_N(a\mid x,\Omega)\\
&\quad+\lambda\log p_P(a\mid x,\Omega).
\end{aligned}
\label{eq:aced-evidence-potential}
\end{equation}
Its negative, $L_{\lambda,\Omega}(a;x)=-s_{\lambda,\Omega}(a;x)$, can be viewed as a weighted evidence cost, with a description-length interpretation up to answer-independent normalization (Appendix~\ref{sec:potential-interpretation}). Stronger fused support thus corresponds to higher potential and lower evidence cost.

\paragraph{Reference-relative information quotient.}
The potential gives a common scale for comparing answers and the replacement decision additionally requires a reference and an acceptance margin. \method{} uses the self-consistency answer $a_0$ as its semantic reference, anchoring the comparison to the natural-language channel that reasons directly over the problem statement. Instantiating the information quotient with $\widetilde q_\lambda$ and $a_0$ gives the log margin
\begin{equation}
\begin{aligned}
\Delta_{\lambda,\Omega}(a;x)
&=\log\frac{\widetilde q_\lambda(a)}
             {\widetilde q_\lambda(a_0)}\\
&=s_{\lambda,\Omega}(a;x)-s_{\lambda,\Omega}(a_0;x)\\
&=\log\frac{n_a+\beta_N}{n_{a_0}+\beta_N}\\
&\quad+\lambda\log\frac{v(a)+\beta_P}{v(a_0)+\beta_P}.
\end{aligned}
\label{eq:aced-reference-margin}
\end{equation}
Common normalizers cancel, leaving a count-level comparison with a separate contribution from each source. A positive margin favors the alternative and the acceptance rule below determines whether that advantage warrants replacement.

\paragraph{Calibrated acceptance.}
If no admissible alternative exists, \method{} returns $a_0$. Otherwise, select the strongest alternative,
\begin{equation}
a_1=\arg\max_{a\in\Omega(x),\,a\ne a_0}
\Delta_{\lambda,\Omega}(a;x),
\label{eq:aced-best-alternative}
\end{equation}
using deterministic tie-breaking. For $\tau\ge0$, the final prediction is
\begin{equation}
\hat a_\Omega(x)=
\begin{cases}
a_1, & \Delta_{\lambda,\Omega}(a_1;x)>\tau,\\
a_0, & \text{otherwise}.
\end{cases}
\label{eq:aced-acceptance}
\end{equation}
The evidence weight $\lambda$ affects answer ranking, whereas $\tau$ controls the margin required for replacement. At $\tau=0$, the rule maximizes fused support, with $a_0$ preferred whenever it ties for the highest score. A positive threshold requires an additional advantage over the reference.

The two parameters serve different roles: $\lambda$ controls the contribution of executable evidence to the fused support, while $\tau$ determines how much stronger an alternative must be than the self-consistency reference before it is accepted. We select $(\lambda,\tau)$ on a held-out calibration set following the procedure in Section~\ref{sec:aced-setup}, and keep them fixed throughout test evaluation. At test time, \method{} applies this fixed decision rule using only the observed natural-language and executable evidence.

\section{Experiments}

We evaluate the two answer domains under a common decision rule: selection within sampled natural-language answers and selection that also permits executable-only answers. Further analyses examine the roles of witness construction and answer aggregation, performance with a stronger reference, sampling variation, and inference cost.
\begin{table*}[t]
\centering
\small
\setlength{\tabcolsep}{3.5pt}
\renewcommand{\arraystretch}{1.12}
\begin{tabularx}{\textwidth}{l*{7}{>{\raggedleft\arraybackslash}X}}
\toprule
\multicolumn{8}{@{}l}{\textbf{(a) Five benchmarks}\quad NL reference: Qwen2.5-0.5B-Instruct} \\
\addlinespace[3pt]
& \multicolumn{5}{c}{Accuracy (\%) $\uparrow$} & \multicolumn{2}{c}{GSM8K overrides} \\
\cmidrule(lr){2-6}\cmidrule(l){7-8}
Method & GSM8K & SVAMP & ASDiv & MultiArith & MAWPS & Fix $\uparrow$ & Damage $\downarrow$ \\
\midrule
\scbase{}@16 (reference) & 31.8 & 54.0 & 59.0 & 97.8 & 84.1 & -- & -- \\
\addlinespace[3pt]
\multicolumn{8}{@{}l}{\textit{Candidate-only selection}} \\
\acr{} & 31.1 & 50.7 & 58.6 & 96.7 & 84.1 & 72 & 81 \\
\prove{} & 49.6 & 68.0 & 71.1 & \textbf{100.0} & 90.7 & 244 & 9 \\
\rowcolor{acedrow}
\textbf{\cand{}@16} & \textbf{58.3} & \textbf{79.3} & \textbf{80.9} & \textbf{100.0} & \textbf{93.2} & \textbf{353} & \textbf{3} \\
\addlinespace[3pt]
\multicolumn{8}{@{}l}{\textit{All-feasible selection}} \\
\psc{}@16 & 82.6 & 90.7 & 84.8 & \textbf{100.0} & 95.3 & 692 & 21 \\
\rowcolor{acedrow}
\textbf{\allaced{}@16} & \textbf{86.9} & \textbf{91.3} & \textbf{86.7} & \textbf{100.0} & \textbf{96.4} & \textbf{733} & \textbf{6} \\
\bottomrule
\end{tabularx}

\vspace{7pt}
\begin{tabularx}{\textwidth}{l*{5}{>{\raggedleft\arraybackslash}X}r}
\toprule
\multicolumn{7}{@{}l}{\textbf{(b) Stronger-reference evaluation}\quad NL reference: Qwen2.5-7B-Instruct} \\
\addlinespace[3pt]
& \multicolumn{2}{c}{GSM8K performance} & \multicolumn{3}{c}{GSM8K overrides} & \\
\cmidrule(lr){2-3}\cmidrule(lr){4-6}
Method & Acc. (\%) $\uparrow$ & Correct $\uparrow$ & Fix $\uparrow$ & Damage $\downarrow$ & Total & $p$ vs. Strong-SC \\
\midrule
Strong-SC@16 (reference) & 79.8 & 1053 & -- & -- & 0 & -- \\
Candidate oracle & 92.2 & 1216 & 163 & 0 & -- & -- \\
\midrule
\psc{}@16 & 82.6 & 1090 & 129 & 92 & 327 & 0.0153 \\
\psc{}@32 & 83.2 & 1098 & 136 & 91 & 329 & 0.00341 \\
\rowcolor{acedrow}
\textbf{\cand{}@16} & 87.3 & 1152 & 117 & \textbf{18} & 165 & $5.72{\cdot}10^{-19}$ \\
\rowcolor{acedrow}
\textbf{\allaced{}@16} & \textbf{88.6} & \textbf{1169} & \textbf{152} & 36 & 255 & $3.67{\cdot}10^{-18}$ \\
\bottomrule
\end{tabularx}
\caption{Results across five benchmarks with a 0.5B natural-language reference in (a) and on GSM8K with a stronger 7B reference in (b). Fix and Damage are measured relative to the corresponding SC reference,
and $p$-values in (b) are obtained from exact paired McNemar tests against Strong-SC. Shaded rows denote \method{}. Bold indicates the best accuracy, correct count, Fix, and Damage within each answer domain in (a) and across selectors in (b), excluding reference and oracle rows.}
\label{tab:main-results}
\end{table*}

\subsection{Experimental Setup and Calibration}
\label{sec:aced-setup}

\paragraph{Benchmarks.}
We evaluate on GSM8K \citep{cobbe2021gsm8k}, SVAMP \citep{patel2021svamp}, ASDiv \citep{miao2020asdiv}, MultiArith \citep{roy2015multiarith}, and MAWPS \citep{koncel2016mawps}, with 1,319, 150, 256, 90, and 961 test examples respectively.

\paragraph{Models and inference budgets.}
Qwen2.5-0.5B-Instruct generates natural-language solutions, while Qwen2.5-Coder-7B-Instruct generates executable programs \citep{yang2024qwen25,hui2024qwen25coder}. The main experiments sample $N=16$ solutions and $M=16$ programs per problem. Additional experiments vary the program budget over $M\in\{8,32\}$ for \method{} and up to $M=64$ for direct program aggregation. All methods use the same answer extraction and normalization interface, with generation and filtering details provided in Appendix~\ref{sec:reproducibility}.

\paragraph{Baselines.}
Comparisons follow the two answer domains defined above. Within the sampled natural-language answer set, \scbase{} returns the majority answer, \acr{} re-scores candidates with an auxiliary judge \citep{jeong2025acr}, and \prove{} verifies candidates by translating their reasoning into programs \citep{toh2024prove}. When executable-only answers are allowed, \psc{} generates programs from the problem using its direct-solver prompt and returns the majority output \citep{chen2023programofthoughts}.  Table~\ref{tab:ablation} compare executable voting with the \method{} selector on the same witness pool.

\paragraph{Evaluation protocol.}
We report accuracy on all five benchmarks. On GSM8K, we also measure \emph{Fix}, which counts incorrect SC references corrected by the selector, and \emph{Damage}, which counts correct references changed to incorrect answers. For an evaluation set $\mathcal D$,
\begin{align}
\mathrm{Damage}_{\mathrm{SC}}
&=\sum_{x\in\mathcal D}\mathbf{1}
\left\{\begin{gathered}
\hat a_\Omega(x)\ne a_0(x)\\
\land\;a_0(x)=a^\star
\end{gathered}\right\},
\label{eq:aced-damage}\\
\mathrm{Fix}
&=\sum_{x\in\mathcal D}\mathbf{1}
\left\{\begin{gathered}
\hat a_\Omega(x)=a^\star\\
\land\;a_0(x)\ne a^\star
\end{gathered}\right\}.
\label{eq:aced-fix}
\end{align}
For candidate-only selection, oracle-gap utilization (OGU) measures the fraction of the accuracy gap between SC and the candidate oracle recovered by the selector:
\begin{equation}
\mathrm{OGU}
=\frac{\mathrm{Acc}(\hat a_{\Omega_{\mathrm{cand}}})
       -\mathrm{Acc}(a_{\mathrm{SC}})}
      {\mathrm{Acc}(\mathrm{Oracle}_{\Omega_{\mathrm{cand}}})
       -\mathrm{Acc}(a_{\mathrm{SC}})}.
\label{eq:aced-ogu}
\end{equation}
The oracle counts a problem as solved whenever a correct answer appears among the sampled natural-language solutions. We use exact paired McNemar tests for comparisons on the same examples. Cost comparisons report total input and output tokens, token counts weighted by model size as a computational proxy, and generation plus execution time measured under the same setup.

\paragraph{Calibration.}
We calibrate $(\lambda,\tau)$ on 500 GSM8K training examples comprising 350 conflict and 150 non-conflict cases, selected with seed 13. For each answer domain and program budget, we evaluate 48 parameter
settings and select one with the highest accuracy among those with at most 10 damaging overrides. All evaluated settings satisfy this cap, and ties are resolved using the fixed representatives in
Appendix~\ref{sec:protocol}, which also reports the complete grid. The selected parameters remain fixed across all five test sets, with $(\lambda,\tau)=(4,0)$ used in both domains at $M=16$.

\begin{figure*}[!t]
\centering
\includegraphics[width=\textwidth]{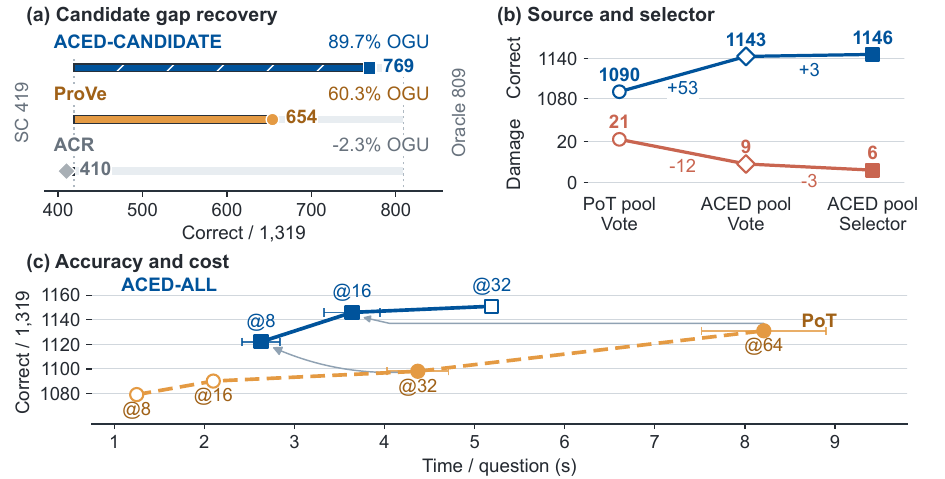}
\caption{GSM8K selection and inference cost.
(a) Oracle-gap utilization for candidate selection.
(b) Witness construction and selector contributions at $M=16$, with the last two configurations sharing witnesses.
(c) Correct predictions versus wall time, with program budgets labeled and arrows marking the reported cost comparisons. Filled markers and horizontal bars show mean time and one standard deviation over three runs, while open markers show single measurements. Correct counts come from the reported evaluation, not averages over timing runs.}
\label{fig:cost}
\end{figure*}

\subsection{Results Across Benchmarks}

Table~\ref{tab:main-results}(a) first evaluates selection without expanding
the answer set. \cand{} achieves the best or tied-best accuracy on all
five benchmarks and exceeds \prove{} by 2.5 to 11.3 percentage points
on the four non-tied benchmarks. Because every eligible answer already
appears in the natural-language samples, these gains isolate better
selection rather than increased answer coverage. Problem-only
witnesses therefore help recover correct minority answers that the
natural-language vote does not select.

The all-feasible block evaluates the second role of executable
evidence. \allaced{} improves over direct \psc{} on four benchmarks
and matches the saturated result on MultiArith. On GSM8K, it produces
41 additional fixes and 15 fewer damaging overrides, yielding 56
additional correct predictions. The improvement combines correction
of SC errors with preservation of initially correct answers. The
following GSM8K analyses examine how candidate selection, witness
construction, and answer aggregation contribute to these gains.

\begin{table*}[!t]
\centering
\small
\setlength{\tabcolsep}{3.5pt}
\renewcommand{\arraystretch}{1.12}
\begin{tabularx}{\textwidth}{l*{7}{>{\raggedleft\arraybackslash}X}}
\toprule
\multirow{2}{*}{Method} & \multirow{2}{*}{Correct $\uparrow$}
& \multicolumn{2}{c}{Tokens / question $\downarrow$}
& \multicolumn{3}{c}{Time / question (s) $\downarrow$}
& \multirow{2}{*}{Speedup $\uparrow$} \\
\cmidrule(lr){3-4}\cmidrule(lr){5-7}
& & Exact & Weighted & Generation & Execution & Wall & \\
\midrule
\multicolumn{8}{@{}l}{\textit{Lower-cost comparison}} \\
\psc{}@32 & 1098 & 7631.2 & 53418.1 & 3.93$\pm$0.35 & 0.44$\pm$0.01 & 4.37$\pm$0.34 & 1.00$\times$ \\
\rowcolor{acedrow}
\textbf{\allaced{}@8} & \textbf{1122} & \textbf{6187.9} & \textbf{21241.0} & \textbf{2.51$\pm$0.22} & \textbf{0.12$\pm$0.01} & \textbf{2.63$\pm$0.21} & \textbf{1.66}$\times$ \\
\midrule
\multicolumn{8}{@{}l}{\textit{Higher-accuracy comparison}} \\
\psc{}@64 & 1131 & 15931.1 & 111517.4 & 7.34$\pm$0.71 & 0.88$\pm$0.02 & 8.21$\pm$0.69 & 1.00$\times$ \\
\rowcolor{acedrow}
\textbf{\allaced{}@16} & \textbf{1146} & \textbf{8979.1} & \textbf{40779.1} & \textbf{3.42$\pm$0.31} & \textbf{0.22$\pm$0.00} & \textbf{3.64$\pm$0.31} & \textbf{2.26}$\times$ \\
\bottomrule
\end{tabularx}
\caption{GSM8K accuracy and inference cost. Token counts include input and output, with model-size weights of 0.5 for natural-language tokens and 7 for code tokens in the weighted column. Timing is mean$\pm$std over seeds 13, 21, and 42 on the same setup. Speedup divides the mean \psc{} wall time by the method's mean wall time within each block.}
\label{tab:main-cost}
\end{table*}
\subsection{Override Analysis}

Figure~\ref{fig:cost}(a) places candidate selection relative to the
SC reference and candidate oracle, which solve 419 and 809 GSM8K
examples, respectively. \cand{} recovers 350 of the 390 available
additional correct predictions, or 89.7\% OGU, compared with 60.3\%
for \prove{}. It also reduces damaging overrides from nine to three.
Thus, executable evidence recovers most of the attainable improvement
within the sampled answer set while preserving more correct SC answers.

In the all-feasible domain, matched-budget
McNemar tests favor \allaced{} over \psc{} at $M=8,16,32$ (all
$p\leq 8.11{\cdot}10^{-4}$), so the main GSM8K improvement is stable
across the evaluated witness budgets. These tests compare the complete
systems, with direct \psc{} using its direct-solver prompt. The
decomposition below separates the witness-pool change from the
same-witness selector comparison.

\subsection{Source--Selector Decomposition}

\begin{table}[tbp]
\centering
\footnotesize
\setlength{\tabcolsep}{3pt}
\begin{tabularx}{\linewidth}{@{}>{\raggedright\arraybackslash}Xrrr@{}}
\toprule
Configuration & Correct & Fix & Damage \\
\midrule
Direct \psc{} pool + executable majority & 1090 & 692 & 21 \\
\addlinespace[3pt]
ACED witness pool + executable majority & 1143 & \textbf{733} & 9 \\
\addlinespace[3pt]
ACED witness pool + ACED selector & \textbf{1146} & \textbf{733} & \textbf{6} \\
\bottomrule
\end{tabularx}
\caption{GSM8K source--selector decomposition at $M=16$. The first two rows hold executable-majority selection fixed while comparing separately generated problem-only witness pools. The last two rows reuse exactly the same ACED witnesses and therefore isolate the effect of the ACED selector.}
\label{tab:ablation}
\end{table}

Figure~\ref{fig:cost}(b) and Table~\ref{tab:ablation} distinguish the
contributions of witness generation and answer selection. Under the
same executable-majority rule, replacing the direct-\psc{} pool with
the ACED witness pool yields 41 additional fixes and 12 fewer damaging
overrides. This accounts for 53 of the 56 additional correct predictions
in the complete-system comparison.

With the ACED witnesses held fixed, the ACED selector retains the
aggregate count of 733 fixes and reduces Damage from nine to six,
adding three correct predictions. The two comparisons identify distinct
contributions. The ACED witness pool improves both correction and
preservation of correct references, while the selector further reduces
harmful replacements on the same evidence. The complete-system gain
therefore reflects both witness construction and answer selection.

\subsection{Robustness to a Stronger Reference and Sampling Variation}

Table~\ref{tab:main-results}(b) evaluates ACED with
Qwen2.5-7B-Instruct as a substantially stronger SC reference. This
reference solves 1053 of 1319 examples, leaving a much narrower
correction set than the main 0.5B model. \allaced{} still reaches 1169
correct predictions. Relative to \psc{}@16, it makes 23 more fixes
and 56 fewer damaging overrides, for a net advantage of 79 examples.
The paired comparison between \allaced{} and \cand{} also favors
admitting executable-only answers, with 39 examples solved only by
\allaced{} and 22 only by \cand{} ($p=0.0396$). The same mechanism
therefore remains effective when the semantic reference is already
strong.

\begingroup
\clubpenalty=10000
Independent witness generation produces the same matched-budget
advantage. Across seeds 3, 13, and 21, \allaced{}@16 averages
$1145.0\pm3.6$ correct predictions, compared with
$1089.0\pm1.0$ for \psc{}@16. The 56-example mean gap closely matches
the main run.
\par\endgroup

Smoothing changes the strength of the all-feasible correction, but
the damage control remains stable. Recalibration over
$\beta_N,\beta_P\in\{0.5,1,2\}$ leaves \cand{} at 769 correct
predictions with 3 damaging overrides. \allaced{} remains between
1134 and 1146 correct with 6 damaging overrides. Complete seed and
smoothing results are reported in the appendices.

\subsection{Inference Cost and Wall-Clock Time}

Figure~\ref{fig:cost}(c) compares correct predictions with wall-clock
time, and Table~\ref{tab:main-cost} provides the corresponding token
costs. \allaced{}@8 solves 24 more examples than \psc{}@32 while
using 18.9\% fewer exact tokens, 60.2\% fewer model-size-weighted
tokens, and 39.8\% less recorded wall time. At the higher-accuracy
operating point, \allaced{}@16 exceeds \psc{}@64 by 15 correct
predictions while using 43.6\% fewer exact tokens and 55.7\% less wall
time. In both comparisons, the complete ACED system is more accurate
with a smaller program budget and lower measured cost.

Timing includes generation in all active branches and sandbox
execution, but excludes model loading. The horizontal error bars in
Figure~\ref{fig:cost}(c) describe timing variation, not variation in
correct counts. Exact tokens provide the hardware-independent
accounting measure, while weighted tokens and same-setup time offer
complementary cost comparisons.

\FloatBarrier
\section{Discussion}

\method{} combines natural-language and executable evidence through KL-based pooling and calibrated reference-relative acceptance. Across five math word-problem benchmarks, it improves candidate selection and the admission of executable-only answers, while controlled analyses show that retaining the SC reference reduces damaging replacements. Our audit of the six damaging \allaced{}@16 overrides on GSM8K identifies semantic-target errors in the executable witnesses.

\section*{Limitations}

The evaluation covers English scalar-answer math problems and Qwen2.5 models, including a stronger 7B natural-language reference. Generalization to open-ended answers and other model families remains untested. Calibration uses a conflict-enriched GSM8K training subset, so the selected operating points need not generalize to other problem distributions. Executable agreement does not verify that programs compute the requested quantity. Wall-clock results characterize the recorded setup rather than hardware-independent speedups.

\bibliography{references}

\clearpage
\appendix
\raggedbottom
\setcounter{topnumber}{3}
\setcounter{bottomnumber}{2}
\setcounter{totalnumber}{5}
\setcounter{dbltopnumber}{3}
\renewcommand{\topfraction}{0.9}
\renewcommand{\bottomfraction}{0.8}
\renewcommand{\textfraction}{0.08}
\renewcommand{\floatpagefraction}{0.7}
\renewcommand{\dbltopfraction}{0.9}
\renewcommand{\dblfloatpagefraction}{0.7}
\makeatletter
\setlength{\@fptop}{0pt}
\setlength{\@fpsep}{12pt}
\setlength{\@fpbot}{0pt plus 1fil}
\setlength{\@dblfptop}{0pt}
\setlength{\@dblfpsep}{12pt}
\setlength{\@dblfpbot}{0pt plus 1fil}
\makeatother
\section{Appendix Overview}
\label{sec:supp-roadmap}

The appendices develop the derivation and experimental evidence behind
the reference-relative decision rule.
\begin{itemize}
\setlength{\itemsep}{2pt}
\item Appendix~\ref{sec:derivations} derives the normalized supports,
KL pool, reference-relative margin, and acceptance rule, and gives the
complete selection algorithm.
\item Appendix~\ref{sec:protocol} records the calibration settings,
sensitivity analyses, and complete parameter grid.
\item Appendix~\ref{sec:additional-results} provides exact benchmark
counts, paired comparisons, same-witness controls, stronger-reference
results, sampling variation, and inference costs.
\item Appendix~\ref{sec:cases} illustrates selection outcomes and
examines all six damaging GSM8K overrides at $M=16$.
\item Appendix~\ref{sec:reproducibility} specifies generation settings,
prompts, baseline configurations, normalization, and output accounting.
\end{itemize}

\section{Evidence-Pooling Derivation}
\label{sec:derivations}

We derive the source distributions, logarithmic pool, and count-level
margin in Section~\ref{sec:aced-method}, then connect the acceptance
rule to its decision interpretation and complete algorithm.

\subsection{Evidence Representation}

The natural-language trajectories and retained program executions are
\begin{align}
\mathcal{E}_{N}(x)
&=\{(y_i,a_i)\}_{i=1}^{N},
\notag\\
n_a
&=|\{i:a_i=a\}|,\\
\mathcal{E}_{P}^{\mathrm{valid}}(x)
&=\{(P_m,o_m)\}_{m=1}^{M_{\mathrm{valid}}},
\notag\\
v(a)
&=\sum_{m=1}^{M_{\mathrm{valid}}}\mathbf{1}[o_m=a].
\end{align}
The SC reference $a_0=\arg\max_{a\in\anl(x)}n_a$ uses a fixed
deterministic tie-breaking order. Each program observes only the
problem, not the natural-language candidates, reasoning traces, SC
answer, or gold label. Executable rows enter
$\mathcal{E}_{P}^{\mathrm{valid}}$ after the label-free safety and
validation checks in Appendix~\ref{sec:normalization-filtering}.
Here $M_{\mathrm{valid}}$ counts retained rows, including those with no
normalized numeric answer.

\subsection{Normalized Supports on the Augmented Space}
\label{sec:augmented-supports}

A retained executable row without a normalized numeric answer is
represented by the non-selectable symbol $\varnothing$. In the
candidate-only domain, numeric outputs outside $\ocand(x)$ are
collected in a second non-selectable bucket $\bot$. Their counts are
\begin{align}
v_{\varnothing}
&=\sum_{m=1}^{M_{\mathrm{valid}}}
\mathbf{1}[o_m=\varnothing],\\
v_{\bot}
&=\sum_{b\in\ap(x)\setminus\ocand(x)}v(b).
\end{align}
The augmented support space used in Eq.~\eqref{eq:aced-kl-pool} is
\begin{equation}
S_\Omega=
\begin{cases}
\ocand(x)\cup\{\bot,\varnothing\},
    & \Omega=\ocand,\\
\oall(x)\cup\{\varnothing\},
    & \Omega=\oall.
\end{cases}
\label{eq:augmented-space}
\end{equation}
For an admissible answer $a$, set $n_a^+=n_a$ and $v_a^+=v(a)$.
Auxiliary buckets have zero natural-language count and the executable
counts defined above. Writing $K_+=|S_\Omega|$, we obtain
\begin{align}
p_N^+(c)
&=\frac{n_c^++\beta_N}{N+\beta_NK_+},\\
p_P^+(c)
&=\frac{v_c^++\beta_P}
{M_{\mathrm{valid}}+\beta_PK_+},
\qquad c\in S_\Omega.
\label{eq:augmented-supports}
\end{align}
Every retained executable row belongs to exactly one selectable answer
or auxiliary bucket. Thus $\sum_c n_c^+=N$ and
$\sum_c v_c^+=M_{\mathrm{valid}}$, so both distributions are normalized.
Positive smoothing makes every entry strictly positive.

For any admissible answers $a,b\in\Omega(x)$, these distributions satisfy
\begin{align}
\frac{p_N^+(a)}{p_N^+(b)}
&=\frac{n_a+\beta_N}{n_b+\beta_N}
=\frac{p_N(a)}{p_N(b)},\\
\frac{p_P^+(a)}{p_P^+(b)}
&=\frac{v(a)+\beta_P}{v(b)+\beta_P}
=\frac{p_P(a)}{p_P(b)}.
\label{eq:augmented-ratios}
\end{align}
Here and below, $p_N(a)$ and $p_P(a)$ abbreviate the selectable-answer
supports in Eqs.~\eqref{eq:aced-support-n}--\eqref{eq:aced-support-p}.
The augmented distributions satisfy $p_N^+(a)=\kappa_N p_N(a)$ and
$p_P^+(a)=\kappa_P p_P(a)$ on the selectable domain, where
\begin{equation}
\begin{aligned}
\kappa_N&=\frac{N+\beta_N|\Omega(x)|}{N+\beta_NK_+},\\
\kappa_P&=
\frac{M_{\mathrm{valid}}+\beta_P|\Omega(x)|}
{M_{\mathrm{valid}}+\beta_PK_+}.
\end{aligned}
\label{eq:normalizer-scaling}
\end{equation}
These constants do not depend on $a$. They cancel from every
admissible-answer ratio, while the auxiliary buckets provide the
normalization required by the KL objective. The selector never
outputs either bucket.

\subsection{Closed-Form KL Pool}
\label{sec:kl-derivation}

We retain the notation $p_N^+,p_P^+$ for the normalized distributions
on $S_\Omega$ throughout the derivation, distinguishing them from the
selectable-answer supports. For $\lambda\ge0$, the main-text objective is
\begin{equation}
J(q)=
\mathrm{KL}(q\Vert p_N^+)
+\lambda\mathrm{KL}(q\Vert p_P^+).
\end{equation}
Expanding the two divergences gives
\begin{equation}
\begin{aligned}
J(q)
&=(1+\lambda)\sum_{c\in S_\Omega}q(c)\log q(c)\\
&\quad-\sum_{c\in S_\Omega}q(c)\log p_N^+(c)\\
&\quad-\lambda\sum_{c\in S_\Omega}q(c)\log p_P^+(c).
\end{aligned}
\end{equation}
Introduce a multiplier $\xi$ for $\sum_c q(c)=1$. The Lagrangian and
its stationarity condition are
\begin{align}
\mathcal{L}(q,\xi)
&=J(q)+\xi\left(\sum_{c\in S_\Omega}q(c)-1\right),\\
\frac{\partial\mathcal{L}}{\partial q(c)}
&=(1+\lambda)(\log q(c)+1)
\notag\\
&\quad-\log p_N^+(c)-\lambda\log p_P^+(c)
\notag\\
&\quad+\xi=0.
\end{align}
Rearranging and normalizing yields
\begin{align}
q_\lambda(c)
&=\frac{
p_N^+(c)^{\frac{1}{1+\lambda}}
p_P^+(c)^{\frac{\lambda}{1+\lambda}}
}{Z_\lambda},\\
Z_\lambda
&=\sum_{d\in S_\Omega}
p_N^+(d)^{\frac{1}{1+\lambda}}
p_P^+(d)^{\frac{\lambda}{1+\lambda}}.
\label{eq:normalized-kl-pool}
\end{align}
The stationary point is strictly positive and $J$ is strictly convex,
so it is the unique minimizer. The result is a logarithmic, or
geometric, pool of the source distributions.

For admissible $a$, substituting
Eq.~\eqref{eq:normalizer-scaling} absorbs $\kappa_N$ and $\kappa_P$
into an answer-independent factor. Consequently,
$q_\lambda(a)\propto
p_N(a)^{1/(1+\lambda)}p_P(a)^{\lambda/(1+\lambda)}$,
as written in Eq.~\eqref{eq:aced-logarithmic-pool}.
This pool supplies the fused support used in the reference comparison.

\subsection{Evidence Potential and Threshold Scale}
\label{sec:potential-interpretation}

\paragraph{Why take logarithms.}
The unscaled support is
$\widetilde q_\lambda(a)=p_N(a)p_P(a)^\lambda$.
The logarithm preserves the ordering of positive supports and converts
their product into an additive contribution from each source:
\begin{equation}
\begin{aligned}
s_{\lambda,\Omega}(a;x)
&=\log p_N(a)+\lambda\log p_P(a),\\
L_{\lambda,\Omega}(a;x)
&=-\log p_N(a)-\lambda\log p_P(a)\\
&=-s_{\lambda,\Omega}(a;x).
\end{aligned}
\label{eq:potential-and-cost}
\end{equation}
All logarithms are natural. The evidence cost is a weighted sum of
negative log-supports. For normalized source distributions, these
terms measure surprisal in nats. On the augmented distributions,
$-\log p_N^+(a)-\lambda\log p_P^+(a)$ differs from this cost by the
answer-independent constant $-\log\kappa_N-\lambda\log\kappa_P$.
The description-length interpretation therefore applies to relative
evidence costs, which agree under either representation.

\paragraph{Scaling the threshold.}
For admissible $a,b\in\Omega(x)$, the normalized KL pool and the
unscaled support obey
\begin{equation}
\log\frac{q_\lambda(a)}{q_\lambda(b)}
=
\frac{1}{1+\lambda}
\log\frac{\widetilde q_\lambda(a)}
{\widetilde q_\lambda(b)}.
\label{eq:pool-ratio-scale}
\end{equation}
A threshold $\bar\tau$ on the left therefore corresponds exactly to
$\tau=(1+\lambda)\bar\tau$ on the unscaled log ratio. For each fixed
$\lambda$, this rescaling preserves the continuous family of threshold
rules. ACED calibrates $\tau$ directly on the unscaled evidence
potential, so its numerical value refers to that scale.

\subsection{Reference-Relative Count Margin}
\label{sec:count-margin}

For admissible answers $a,b\in\Omega(x)$,
Eq.~\eqref{eq:augmented-ratios} connects the normalized distributions
to the support used by the selector:
\begin{equation}
\begin{aligned}
\log\frac{\widetilde q_\lambda(a)}{\widetilde q_\lambda(b)}
&=\log\frac{p_N^+(a)}{p_N^+(b)}
 +\lambda\log\frac{p_P^+(a)}{p_P^+(b)}\\
&=\log\frac{n_a+\beta_N}{n_b+\beta_N}\\
&\quad+\lambda\log\frac{v(a)+\beta_P}{v(b)+\beta_P}.
\end{aligned}
\end{equation}
Instantiating the information quotient with $\widetilde q_\lambda$
and the reference $a_0$ gives
\begin{equation}
\begin{aligned}
\mathcal{Q}_{\lambda,\Omega}(a\succ a_0;x)
&=\frac{\widetilde q_\lambda(a)}
        {\widetilde q_\lambda(a_0)}\\
&=\frac{p_N(a)}{p_N(a_0)}
\left(\frac{p_P(a)}{p_P(a_0)}\right)^\lambda.
\end{aligned}
\end{equation}
Taking the logarithm and setting $b=a_0$ in the general expansion
above gives the potential difference and count-level margin in
Eq.~\eqref{eq:aced-reference-margin}.
For unit smoothing, the implemented statistic reduces to
\begin{equation}
\begin{aligned}
\Delta_{\lambda,\Omega}(a;x)
&=\log\frac{n_a+1}{n_{a_0}+1}\\
&\quad+\lambda\log\frac{v(a)+1}{v(a_0)+1}.
\end{aligned}
\end{equation}
Common normalizers and auxiliary-bucket mass cancel from these ratios,
leaving a separate contribution from each source.
By Eq.~\eqref{eq:potential-and-cost},
\begin{equation}
\Delta_{\lambda,\Omega}(a;x)
=L_{\lambda,\Omega}(a_0;x)-L_{\lambda,\Omega}(a;x).
\end{equation}
A positive margin therefore measures the reduction in weighted
evidence cost from accepting $a$ instead of $a_0$.

\subsection{Calibrated Acceptance and Its Boundary}
\label{sec:zero-threshold}

\paragraph{Parameter roles.}
The weight $\lambda$ changes the fused potential and can change the
ordering of answers. The threshold $\tau$ leaves that ordering
unchanged and controls the margin required to replace $a_0$.
If no admissible alternative exists, the output is $a_0$.
Otherwise, let $a_1$ be the strongest alternative with the fixed
deterministic tie-breaking order.

At $\tau=0$, writing $s(a)=s_{\lambda,\Omega}(a;x)$,
Eq.~\eqref{eq:aced-acceptance} becomes
\begin{equation}
\hat a_\Omega^{(0)}(x)=
\begin{cases}
a_1, & s(a_1)>s(a_0),\\
a_0, & \text{otherwise}.
\end{cases}
\end{equation}
This is global fused-score maximization with $a_0$ retained whenever
it ties for the largest score. A positive $\tau$ adds a replacement
margin beyond that ranking rule. The frozen $M=16$ operating point
uses $(\lambda,\tau)=(4,0)$ in both domains. The all-feasible points
at $M=8$ and $M=32$ use $(4,1)$, adding a positive replacement margin.

\paragraph{Ideal posterior comparison.}
Let $A^\star$ denote the unknown correct-answer random variable and
write
\begin{equation}
\pi(a)=
\Pr(A^\star=a\mid x,\mathcal{E}_N,
                  \mathcal{E}_P^{\mathrm{valid}}).
\end{equation}
For positive posterior support, the ideal relative comparison would be
\begin{equation}
\begin{aligned}
\Delta^\star(a;x)
&=\log\frac{\pi(a)}{\pi(a_0)}\\
&=L^\star(a_0;x)-L^\star(a;x),\\
L^\star(a;x)&=-\log\pi(a).
\end{aligned}
\end{equation}
ACED preserves this relative-comparison form using support constructed
from observed counts. Its evidence potential is a computable surrogate
for this comparison, rather than an estimate of $\pi$.

\paragraph{Cost-sensitive interpretation.}
Consider a local comparison between $a$ and $a_0$, conditioned on one
of them being correct. Let $C_{\mathrm{FA}}>0$ be the cost of accepting
$a$ when $a_0$ is correct, and let $C_{\mathrm{MISS}}>0$ be the cost of
retaining $a_0$ when $a$ is correct. The conditional risks are
\begin{align}
R(\text{accept }a)&=C_{\mathrm{FA}}\,\bar\pi(a_0),\\
R(\text{retain }a_0)&=C_{\mathrm{MISS}}\,\bar\pi(a),
\end{align}
where $\bar\pi$ is $\pi$ renormalized over $\{a,a_0\}$. Accepting $a$
has lower risk when
\begin{equation}
\log\frac{\pi(a)}{\pi(a_0)}
>
\log\frac{C_{\mathrm{FA}}}{C_{\mathrm{MISS}}}.
\end{equation}
This conditional risk comparison motivates the log-ratio threshold.
In ACED, the cost ratio provides an interpretation of the rule, while
held-out accuracy and Damage determine its empirical operating point.
The selector uses pooled support in place of posterior odds and freezes
$(\lambda,\tau)$ before testing, as specified in
Appendix~\ref{sec:protocol}.

\subsection{Complete Selection Algorithm}
\label{sec:complete-algorithm}

Algorithm~\ref{alg:aced} collects evidence construction, pooling, and
acceptance in one procedure. It uses the same normalized-answer
interface and deterministic tie order in both domains. Auxiliary
buckets enter the derivation but are never candidates for selection.

\begin{algorithm}[tbp]
\caption{Information pooling and calibrated acceptance}
\label{alg:aced}
\footnotesize
\begin{tabular}{@{}L{0.97\columnwidth}@{}}
\textbf{Input:} problem $x$; budgets $N,M$; domain
$\Omega\in\{\ocand,\oall\}$; smoothing
$\beta_N,\beta_P>0$; frozen $\lambda,\tau\ge0$.\\
\textbf{Output:} selected answer $\hat a_\Omega(x)$.
\end{tabular}
\vspace{2pt}

\begin{tabular}{@{}r@{\hspace{0.55em}}p{0.91\columnwidth}@{}}
1 & Generate $N$ natural-language reasoning paths. Normalize
    final answers and count $n_a$.\\
2 & Set $a_0\leftarrow\arg\max_a n_a$.\\
3 & Generate $M$ problem-only witnesses. Apply safety checks,
    execute permitted programs, then validate and normalize
    outputs to obtain $M_{\mathrm{valid}}$ and $v(a)$.\\
4 & Construct $\Omega$. If $\Omega\setminus\{a_0\}=\emptyset$,
    \textbf{return} $a_0$.\\
5 & Compute $p_N(a\mid x,\Omega)$, $p_P(a\mid x,\Omega)$,
    and $s_{\lambda,\Omega}(a;x)$ for every $a\in\Omega$.\\
6 & Compute $\Delta_{\lambda,\Omega}(a;x)$ for each
    $a\in\Omega\setminus\{a_0\}$.\\
7 & $a_1\leftarrow\arg\max_{a\in\Omega,\,a\ne a_0}
    \Delta_{\lambda,\Omega}(a;x)$.\\
8 & \textbf{if} $\Delta_{\lambda,\Omega}(a_1;x)>\tau$
    \textbf{then return} $a_1$.\\
9 & \textbf{return} $a_0$.
\end{tabular}
\end{algorithm}

\FloatBarrier
\section{Calibration and Sensitivity}
\label{sec:protocol}

Calibration selects an operating point for each answer domain and
program budget. We report the selected settings, examine their
sensitivity to the Damage cap and smoothing, and provide the full grid.

\subsection{Calibration Protocol}

Table~\ref{tab:protocol} specifies the held-out selection procedure.
The 500-example split contains 153 initially correct SC answers.
Each domain and budget is evaluated at 48 parameter settings, with
the selected setting fixed across all five test benchmarks.

\begin{table}[!htbp]
\centering
\captionsetup{skip=4pt}
\footnotesize
\begin{tabularx}{\linewidth}{@{}L{0.24\linewidth}>{\raggedright\arraybackslash}X@{}}
\toprule
Item & Value \\
\midrule
Calibration data & 500 GSM8K training examples, comprising 350 conflict and 150 non-conflict cases, selected with seed 13 and disjoint from the test sets. \\
Evidence budgets & $N=16$, with $M\in\{8,16,32\}$ for the reported settings. \\
Evidence weights & $\lambda\in\{0.25,0.5,1,2,4,8\}$ \\
Thresholds & $\tau\in\{0,0.25,0.5,1,1.5,2,3,4\}$ \\
Selection criterion & Highest correct count among settings with at most 10 damaging overrides, or $2\%$ of the full split. \\
Tie resolution & Fixed representatives in Table~\ref{tab:calibration-audit}, chosen before test evaluation. \\
\bottomrule
\end{tabularx}
\caption{Calibration data, search grid, and operating-point selection.}
\label{tab:protocol}
\end{table}

\subsection{Selected Operating Points}

Table~\ref{tab:calibration-audit} reports the fixed representatives
and their calibration and GSM8K test outcomes. Several settings share
the highest correct count. The reported representatives resolve these
ties before test evaluation. The complete grid appears in
Table~\ref{tab:calibration-grid}.

Conditional on the 153 SC-correct calibration examples, the selected
candidate-only points incur $0/153$ Damage and the all-feasible
points incur $4/153$. The corresponding nominal one-sided binomial
upper bounds are 0.030 and 0.074, using a tail probability of
$0.05/5$ for the five reported points. These descriptive bounds
adjust for the five reported points, but not for selection over the
full calibration grid.

\begin{table*}[!tbp]
\centering
\footnotesize
\setlength{\tabcolsep}{4pt}
\begin{tabular*}{\textwidth}{@{\extracolsep{\fill}}lrr rrrr rrrr@{}}
\toprule
& & & \multicolumn{4}{c}{Calibration ($n=500$)}
& \multicolumn{4}{c}{GSM8K test ($n=1319$)} \\
\cmidrule(lr){4-7}\cmidrule(l){8-11}
Setting & $\lambda$ & $\tau$
& Correct & Fix & Damage & Overrides
& Correct & Fix & Damage & Overrides \\
\midrule
Cand@16 & 4 & 0 & 290 & 137 & 0 & 158 & 769 & 353 & 3 & 421 \\
Cand@32 & 2 & 0 & 291 & 138 & 0 & 161 & 773 & 357 & 3 & 437 \\
\midrule
All@8 & 4 & 1 & 473 & 324 & 4 & 350 & 1122 & 710 & 7 & 841 \\
All@16 & 4 & 0 & 472 & 323 & 4 & 350 & 1146 & 733 & 6 & 864 \\
All@32 & 4 & 1 & 473 & 324 & 4 & 350 & 1151 & 738 & 6 & 867 \\
\bottomrule
\end{tabular*}
\caption{Frozen operating points and their calibration and test outcomes.
Cand and All denote \cand{} and \allaced{}, and the suffix gives $M$.
The calibration and test SC references are correct on 153 and 419
examples, respectively. Overrides include wrong-to-wrong changes
as well as Fix and Damage.}
\label{tab:calibration-audit}
\end{table*}

\begin{figure*}[!tbp]
\centering
\includegraphics[width=0.96\textwidth]{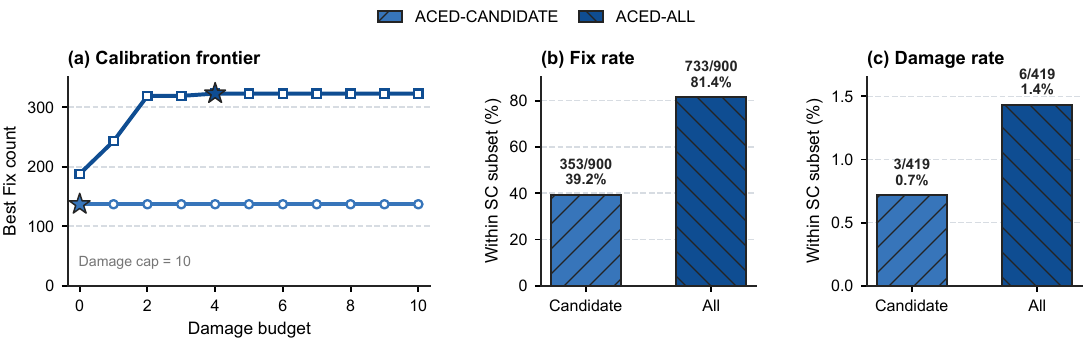}
\caption{Calibration and GSM8K override behavior at $M=16$. (a) Best calibration Fix count under each Damage budget, with stars at the frozen settings. (b) Fix rate among the 900 \scbase{}-wrong test examples. (c) Damage rate among the 419 \scbase{}-correct test examples.}
\label{fig:calibration-frontier}
\end{figure*}

\subsection{Damage-Cap Sensitivity}
The largest Damage count anywhere in the calibration grid is four,
below the limit of 10. Every setting is therefore eligible under the
reported cap, and selection is determined by correct count and the
fixed tie representatives. The $M=16$ points use $\tau=0$, while the
all-feasible points at $M=8$ and $M=32$ use $\tau=1$.

Table~\ref{tab:cap-sensitivity} recomputes the selection rule under
several caps using the same calibration grid. Candidate-only selection
is unchanged because every candidate-domain point on the selected
plateau has zero Damage. In the all-feasible domain, caps from 1\% to
5\% select the same high-accuracy plateau. Requiring zero observed
Damage instead selects points with fewer corrections. This contrast
quantifies the correction--damage trade-off on held-out data.

\begin{table}[!htbp]
\centering
\footnotesize
\setlength{\tabcolsep}{4pt}
\begin{tabular*}{\linewidth}{@{\extracolsep{\fill}}lcc@{}}
\toprule
Calibration setting & 0\% cap & 1--5\% caps \\
\midrule
Cand@16 & 290/137/0 & 290/137/0 \\
Cand@32 & 291/138/0 & 291/138/0 \\
All@8 & 327/174/0 & 473/324/4 \\
All@16 & 341/188/0 & 472/323/4 \\
All@32 & 288/135/0 & 473/324/4 \\
\bottomrule
\end{tabular*}
\caption{Calibration-cap sensitivity on the fixed 500-example split. Each entry is Correct/Fix/Damage for the best eligible point. Caps of 1\%, 2\%, and 5\% select the same plateau.}
\label{tab:cap-sensitivity}
\end{table}

\subsection{Smoothing Sensitivity}
The main paper uses unit Dirichlet smoothing,
$\beta_N=\beta_P=1$. Recalibrating every setting over
$\beta_N,\beta_P\in\{0.5,1,2\}$ preserves the central result:
candidate-only accuracy is essentially unchanged, and all-feasible
Damage remains fixed at 6 for $M\in\{16,32\}$ while Correct stays
within 12 and 7 examples of the main points. Table~\ref{tab:beta-summary}
reports the complete ranges, including the broader variation at $M=8$.

\begin{table}[!htbp]
\centering
\footnotesize
\setlength{\tabcolsep}{3pt}
\begin{tabular*}{\linewidth}{@{\extracolsep{\fill}}lrrr@{}}
\toprule
Domain / $M$ & Correct & Fix & Damage \\
\midrule
Cand / 16 & 769--769 & 353--353 & 3--3 \\
Cand / 32 & 771--773 & 356--357 & 3--4 \\
All / 8 & 1109--1126 & 697--714 & 7--7 \\
All / 16 & 1134--1146 & 721--733 & 6--6 \\
All / 32 & 1144--1151 & 731--738 & 6--6 \\
\bottomrule
\end{tabular*}
\caption{Sensitivity to $\beta_N,\beta_P\in\{0.5,1,2\}$. Each setting recalibrates $\lambda$ and $\tau$ under the same pre-test rule. Entries give the range over the nine smoothing pairs. The unit-smoothing operating points appear in Table~\ref{tab:calibration-audit}.}
\label{tab:beta-summary}
\end{table}

\subsection{Complete Calibration Grid}
\label{sec:full-grid}

Table~\ref{tab:calibration-grid} gives all 48 settings for each of the
five reported domain--budget combinations. Each entry lists
Correct/Fix/Damage on the same 500 calibration examples. The SC
reference is correct on 153 of these examples, so every entry
satisfies $\mathrm{Correct}=153+\mathrm{Fix}-\mathrm{Damage}$.
Bold entries identify the frozen operating points rather than every
setting on a tied maximum.

\begin{table*}[tp]
\centering
\footnotesize
\renewcommand{\arraystretch}{1.04}
\setlength{\tabcolsep}{3pt}
\begin{tabular*}{\textwidth}{@{\extracolsep{\fill}}lrrrrrrrr@{}}
\toprule
& \multicolumn{8}{c}{$\tau$} \\
\cmidrule(l){2-9}
$\lambda$ & 0 & 0.25 & 0.5 & 1 & 1.5 & 2 & 3 & 4 \\
\midrule
\multicolumn{9}{@{}l}{\textit{\cand{}@$M=16$}} \\
0.25 & 269/116/0 & 234/81/0 & 188/35/0 & 153/0/0 & 153/0/0 & 153/0/0 & 153/0/0 & 153/0/0 \\
0.5 & 290/137/0 & 290/137/0 & 290/137/0 & 223/70/0 & 153/0/0 & 153/0/0 & 153/0/0 & 153/0/0 \\
1 & 290/137/0 & 290/137/0 & 290/137/0 & 290/137/0 & 290/137/0 & 274/121/0 & 153/0/0 & 153/0/0 \\
2 & 290/137/0 & 290/137/0 & 290/137/0 & 290/137/0 & 290/137/0 & 290/137/0 & 290/137/0 & 290/137/0 \\
4 & \textbf{290/137/0} & 290/137/0 & 290/137/0 & 290/137/0 & 290/137/0 & 290/137/0 & 290/137/0 & 290/137/0 \\
8 & 290/137/0 & 290/137/0 & 290/137/0 & 290/137/0 & 290/137/0 & 290/137/0 & 290/137/0 & 290/137/0 \\
\addlinespace[2pt]
\multicolumn{9}{@{}l}{\textit{\cand{}@$M=32$}} \\
0.25 & 280/127/0 & 249/96/0 & 209/56/0 & 153/0/0 & 153/0/0 & 153/0/0 & 153/0/0 & 153/0/0 \\
0.5 & 291/138/0 & 291/138/0 & 290/137/0 & 271/118/0 & 193/40/0 & 153/0/0 & 153/0/0 & 153/0/0 \\
1 & 291/138/0 & 291/138/0 & 291/138/0 & 290/137/0 & 290/137/0 & 290/137/0 & 228/75/0 & 153/0/0 \\
2 & \textbf{291/138/0} & 291/138/0 & 291/138/0 & 291/138/0 & 290/137/0 & 290/137/0 & 290/137/0 & 290/137/0 \\
4 & 291/138/0 & 291/138/0 & 291/138/0 & 291/138/0 & 291/138/0 & 291/138/0 & 290/137/0 & 290/137/0 \\
8 & 291/138/0 & 291/138/0 & 291/138/0 & 291/138/0 & 291/138/0 & 291/138/0 & 291/138/0 & 291/138/0 \\
\addlinespace[2pt]
\multicolumn{9}{@{}l}{\textit{\allaced{}@$M=8$}} \\
0.25 & 246/93/0 & 203/50/0 & 182/29/0 & 153/0/0 & 153/0/0 & 153/0/0 & 153/0/0 & 153/0/0 \\
0.5 & 298/145/0 & 284/131/0 & 249/96/0 & 182/29/0 & 153/0/0 & 153/0/0 & 153/0/0 & 153/0/0 \\
1 & 465/314/2 & 455/304/2 & 422/270/1 & 327/174/0 & 277/124/0 & 188/35/0 & 153/0/0 & 153/0/0 \\
2 & 472/323/4 & 472/323/4 & 472/323/4 & 472/323/4 & 472/323/4 & 467/316/2 & 372/220/1 & 206/53/0 \\
4 & 473/324/4 & 473/324/4 & 473/324/4 & \textbf{473/324/4} & 473/324/4 & 473/324/4 & 473/324/4 & 473/324/4 \\
8 & 473/324/4 & 473/324/4 & 473/324/4 & 473/324/4 & 473/324/4 & 473/324/4 & 473/324/4 & 473/324/4 \\
\addlinespace[2pt]
\multicolumn{9}{@{}l}{\textit{\allaced{}@$M=16$}} \\
0.25 & 270/117/0 & 234/81/0 & 188/35/0 & 153/0/0 & 153/0/0 & 153/0/0 & 153/0/0 & 153/0/0 \\
0.5 & 377/225/1 & 334/181/0 & 296/143/0 & 223/70/0 & 153/0/0 & 153/0/0 & 153/0/0 & 153/0/0 \\
1 & 469/319/3 & 470/319/2 & 468/317/2 & 434/283/2 & 341/188/0 & 278/125/0 & 153/0/0 & 153/0/0 \\
2 & 472/323/4 & 472/323/4 & 472/323/4 & 472/323/4 & 472/323/4 & 472/323/4 & 470/319/2 & 395/243/1 \\
4 & \textbf{472/323/4} & 472/323/4 & 472/323/4 & 472/323/4 & 472/323/4 & 472/323/4 & 472/323/4 & 472/323/4 \\
8 & 472/323/4 & 472/323/4 & 472/323/4 & 472/323/4 & 472/323/4 & 472/323/4 & 472/323/4 & 472/323/4 \\
\addlinespace[2pt]
\multicolumn{9}{@{}l}{\textit{\allaced{}@$M=32$}} \\
0.25 & 288/135/0 & 249/96/0 & 209/56/0 & 153/0/0 & 153/0/0 & 153/0/0 & 153/0/0 & 153/0/0 \\
0.5 & 432/280/1 & 399/247/1 & 342/190/1 & 273/120/0 & 193/40/0 & 153/0/0 & 153/0/0 & 153/0/0 \\
1 & 470/321/4 & 470/321/4 & 470/321/4 & 471/319/1 & 452/300/1 & 377/225/1 & 228/75/0 & 153/0/0 \\
2 & 472/323/4 & 472/323/4 & 472/323/4 & 472/323/4 & 472/323/4 & 472/323/4 & 472/323/4 & 472/322/3 \\
4 & 473/324/4 & 473/324/4 & 473/324/4 & \textbf{473/324/4} & 473/324/4 & 473/324/4 & 473/324/4 & 473/324/4 \\
8 & 473/324/4 & 473/324/4 & 473/324/4 & 473/324/4 & 473/324/4 & 473/324/4 & 473/324/4 & 473/324/4 \\
\bottomrule
\end{tabular*}
\caption{Complete calibration sweep for the five reported domain and budget settings. Each cell is Correct/Fix/Damage on 500 GSM8K training examples. Bold marks the frozen representative selected from the highest-correct eligible plateau.}
\label{tab:calibration-grid}
\end{table*}

\section{Additional Experimental Results}
\label{sec:additional-results}

These results supplement the main comparisons with exact benchmark
counts, paired tests, reference-strength and sampling analyses, and
detailed inference costs. Unless stated otherwise, Fix and Damage use
the 0.5B \scbase{}@16 reference.

\subsection{Accuracy Decomposition}
\label{sec:evaluation-counts}

Section~\ref{sec:aced-setup} defines Damage and Fix in
Eqs.~\eqref{eq:aced-damage} and~\eqref{eq:aced-fix}, relative to the
SC reference used in each setting. For an evaluation set $\mathcal D$,
these counts have an exact relation to the change in correct count:
\begin{equation}
\begin{aligned}
&\sum_{x\in\mathcal D}\mathbf{1}\{\hat a_\Omega(x)=a^\star\}\\
&\quad-\sum_{x\in\mathcal D}\mathbf{1}\{a_0(x)=a^\star\}\\
&\qquad=\mathrm{Fix}-\mathrm{Damage}_{SC}.
\end{aligned}
\label{eq:fix-damage-identity}
\end{equation}
Each example contributes $+1$ when a wrong reference is corrected,
$-1$ when a correct reference is replaced, and zero otherwise.
Dividing by $|\mathcal D|$ gives the corresponding accuracy difference.

For candidate-only selection, define the candidate-oracle correct count
as
\begin{equation}
C_{\mathrm{oracle}}
=\sum_{x\in\mathcal D}
\mathbf{1}\{a^\star\in\anl(x)\},
\end{equation}
and let $C_{\mathrm{SC}}$ be the SC correct count. Whenever
$C_{\mathrm{oracle}}>C_{\mathrm{SC}}$,
Eq.~\eqref{eq:aced-ogu} is equivalently
\begin{equation}
\mathrm{OGU}
=\frac{\mathrm{Fix}-\mathrm{Damage}_{SC}}
       {C_{\mathrm{oracle}}-C_{\mathrm{SC}}}.
\end{equation}
Thus OGU measures the net improvement recovered within the fixed
natural-language answer set.

\subsection{Benchmark Results and Paired Comparisons}

\paragraph{Exact benchmark counts.}
Table~\ref{tab:benchmark-results} gives the exact counts underlying
Table~\ref{tab:main-results}(a), together with the $M=32$ all-feasible
results. For each domain and budget, the same frozen operating point
is applied across the five benchmarks.

\begin{table*}[!tbp]
\centering
\captionsetup{skip=4pt}
\footnotesize
\setlength{\tabcolsep}{3pt}
\begin{tabular*}{\textwidth}{@{\extracolsep{\fill}}lrrrrr@{}}
\toprule
Method & GSM8K & SVAMP & ASDiv & MultiArith & MAWPS \\
\midrule
\multicolumn{6}{@{}l}{\textit{Candidate-only selection}} \\
\scbase{}@16 & 419/1319 & 81/150 & 151/256 & 88/90 & 808/961 \\
\acr{} & 410/1319 & 76/150 & 150/256 & 87/90 & 808/961 \\
\prove{} & 654/1319 & 102/150 & 182/256 & \textbf{90/90} & 872/961 \\
\cand{}@16 & \textbf{769/1319} & \textbf{119/150} & \textbf{207/256} & \textbf{90/90} & \textbf{896/961} \\
\midrule
\multicolumn{6}{@{}l}{\textit{All-feasible selection}} \\
\psc{}@16 & 1090/1319 & 136/150 & 217/256 & \textbf{90/90} & 916/961 \\
\allaced{}@16 & \textbf{1146/1319} & \textbf{137/150} & \textbf{222/256} & \textbf{90/90} & \textbf{926/961} \\
\allaced{}@32 & 1151/1319 & 138/150 & 223/256 & 90/90 & 927/961 \\
\bottomrule
\end{tabular*}
\caption{Exact correct counts corresponding to the five-benchmark main table, with the additional \allaced{}@32 row shown for budget scaling. Bold marks the best $M=16$ result within each answer-domain block, including ties.}
\label{tab:benchmark-results}
\end{table*}

\paragraph{Matched-budget paired comparisons.}
Table~\ref{tab:paired-main} reports the paired counts used for the GSM8K significance tests in the main paper. Every row compares the same examples at the same code-sample budget.

\begin{table}[!htbp]
\centering
\footnotesize
\setlength{\tabcolsep}{3pt}
\begin{tabularx}{\linewidth}{@{}>{\raggedright\arraybackslash}Xrrrr@{}}
\toprule
Comparison & A-only & B-only & $\Delta$ & Exact $p$ \\
\midrule
\allaced{}@8 vs. \psc{}@8 & 101 & 58 & +43 & $8.11{\cdot}10^{-4}$ \\
\allaced{}@16 vs. \psc{}@16 & 105 & 49 & +56 & $7.50{\cdot}10^{-6}$ \\
\allaced{}@32 vs. \psc{}@32 & 103 & 50 & +53 & $2.20{\cdot}10^{-5}$ \\
\bottomrule
\end{tabularx}
\caption{Matched-budget GSM8K comparisons. A-only and B-only count examples solved exclusively by the first and second methods. $\Delta$ is their difference, and $p$ is the exact paired McNemar result.}
\label{tab:paired-main}
\end{table}

\begin{figure*}[!tbp]
\centering
\includegraphics[width=0.97\textwidth]{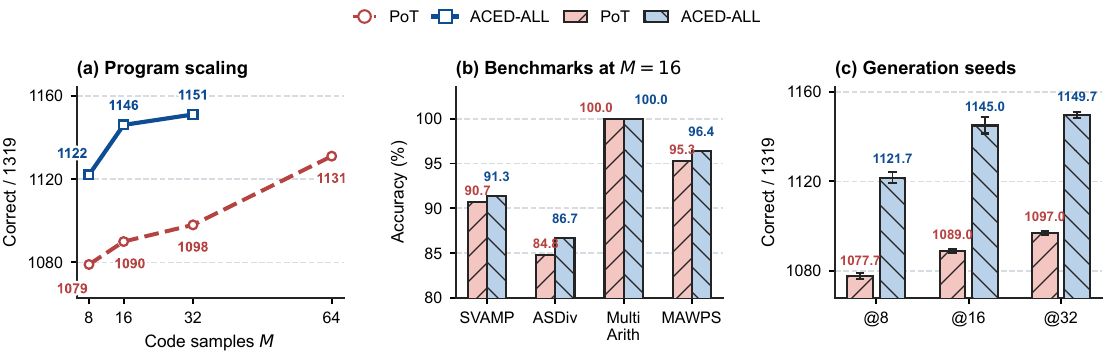}
\caption{Results across code budgets, benchmarks, and generation seeds. (a) GSM8K correct counts as the code budget increases. (b) Accuracy on the four additional benchmarks at $M=16$. (c) GSM8K correct counts over generation seeds 3, 13, and 21.}
\label{fig:appendix-diagnostics}
\end{figure*}

\paragraph{Fix and Damage on additional benchmarks.}
Table~\ref{tab:additional-ledgers} decomposes the accuracy gains on
SVAMP, ASDiv, and MAWPS. On SVAMP, two additional fixes outweigh one
additional damaging override. On ASDiv and MAWPS, \allaced{} improves
both correction and preservation of correct references.

\begin{table}[!htbp]
\centering
\footnotesize
\setlength{\tabcolsep}{5pt}
\begin{tabular*}{\linewidth}{@{\extracolsep{\fill}}llrrr@{}}
\toprule
Dataset & Method & Correct & Fix & Damage \\
\midrule
SVAMP & \psc{}@16 & 136/150 & 58 & \textbf{3} \\
      & \allaced{}@16 & \textbf{137/150} & \textbf{60} & 4 \\
\midrule
ASDiv & \psc{}@16 & 217/256 & 71 & 5 \\
      & \allaced{}@16 & \textbf{222/256} & \textbf{73} & \textbf{2} \\
\midrule
MAWPS & \psc{}@16 & 916/961 & 124 & 16 \\
      & \allaced{}@16 & \textbf{926/961} & \textbf{126} & \textbf{8} \\
\bottomrule
\end{tabular*}
\caption{Correct counts, Fix, and Damage at $M=16$, relative to each dataset's \scbase{}@16 reference. Bold marks the better value within each dataset and metric.}
\label{tab:additional-ledgers}
\end{table}

\subsection{Witness Construction and Answer Aggregation}
\label{sec:source-selector}
We distinguish the complete-system comparison from the same-witness
control. Direct \psc{} generates its own problem-only program pool and
returns its program majority. Applying program-majority selection to
the ACED witness pool gives
\begin{equation}
\hat a_{\mathrm{vote}}(x)
=
\arg\max_{a\in\ap(x)}v_{\mathrm{ACED}}(a),
\end{equation}
with the fixed program-vote tie rule. The ACED selector combines both
source supports through Eq.~\eqref{eq:aced-acceptance}, using the
frozen $(\lambda,\tau)$.
The direct \psc{} comparison includes the distinct problem-only
prompts. In contrast, the executable-majority and ACED rows in
Table~\ref{tab:m-scaling-ablation} reuse the same witness outputs,
isolating the contribution of natural-language support and the
ACED aggregation rule.

Table~\ref{tab:ablation} in the main paper reports the $M=16$
decomposition. With executable-majority selection fixed, changing
from the direct-\psc{} pool to the ACED pool adds 53 correct
predictions. With the ACED pool fixed, using both source supports
adds three more. At this $\tau=0$ operating point, the selector is
logarithmic-pool ranking with ties favoring SC, as derived in
Appendix~\ref{sec:zero-threshold}.

\subsection{Same-Witness Comparisons Across Budgets}
Table~\ref{tab:m-scaling-ablation} extends the fixed-witness comparison
to $M=8,16,32$. ACED reduces Damage at all three budgets. At $M=16$
and $M=32$, the Fix counts remain 733 and 738, respectively, so the
reductions in Damage account for the gains in correct predictions.
At $M=8$, the calibrated $\tau=1$ rule avoids four damaging overrides
at the cost of six fixes, giving two fewer correct predictions on the
fixed pool. The complete-system comparison instead shows ACED
exceeding direct \psc{} by
43 correct predictions at the same budget
(Table~\ref{tab:paired-main}).

\begin{table}[!htbp]
\centering
\footnotesize
\setlength{\tabcolsep}{3pt}
\begin{tabular*}{\linewidth}{@{\extracolsep{\fill}}rlrrr@{}}
\toprule
$M$ & Decision rule & Correct & Fix & Damage \\
\midrule
8 & Executable majority & \textbf{1124} & \textbf{716} & 11 \\
8 & ACED selector & 1122 & 710 & \textbf{7} \\
\midrule
16 & Executable majority & 1143 & \textbf{733} & 9 \\
16 & ACED selector & \textbf{1146} & \textbf{733} & \textbf{6} \\
\midrule
32 & Executable majority & 1147 & \textbf{738} & 10 \\
32 & ACED selector & \textbf{1151} & \textbf{738} & \textbf{6} \\
\bottomrule
\end{tabular*}
\caption{All-feasible decisions on the same ACED witness pool. Executable majority uses only $v_{\mathrm{ACED}}(a)$, while the ACED selector combines natural-language and executable support using the frozen operating point. Bold marks the better value within each budget and metric.}
\label{tab:m-scaling-ablation}
\end{table}

\subsection{Stronger Natural-Language Reference}
\label{sec:strong-reference-details}

Table~\ref{tab:main-results}(b) gives the complete correct, Fix, Damage,
and override counts with Qwen2.5-7B-Instruct as the reference.
Table~\ref{tab:strong-reference-paired} adds the paired counts used
by the significance tests. Here, Fix and Damage are measured against
the 1053 correct and 266 incorrect Strong-SC predictions.
The comparison between \allaced{} and \cand{} tests whether admitting
executable-only answers remains useful with this stronger reference.

\begin{table}[!htbp]
\centering
\footnotesize
\setlength{\tabcolsep}{3pt}
\begin{tabularx}{\linewidth}{@{}>{\raggedright\arraybackslash}Xrrr@{}}
\toprule
Comparison & A-only & B-only & McNemar $p$ \\
\midrule
\cand{}@16 vs. Strong-SC@16 & 117 & 18 & $5.72{\cdot}10^{-19}$ \\
\allaced{}@16 vs. Strong-SC@16 & 152 & 36 & $3.67{\cdot}10^{-18}$ \\
\allaced{}@16 vs. \cand{}@16 & 39 & 22 & 0.0396 \\
\psc{}@16 vs. Strong-SC@16 & 129 & 92 & 0.0153 \\
\psc{}@32 vs. Strong-SC@16 & 136 & 91 & 0.00341 \\
\bottomrule
\end{tabularx}
\caption{Paired GSM8K counts with the stronger SC reference. A-only and B-only are examples solved only by the first or second method, respectively.}
\label{tab:strong-reference-paired}
\end{table}

\subsection{Variation across Generation Seeds}
\label{sec:seed-details}
Table~\ref{tab:seed} reports GSM8K results for generation seeds 3, 13,
and 21. \allaced{} exceeds \psc{} at each matched budget in every
run. These generation runs measure variation in answer selection,
whereas the repetitions in Table~\ref{tab:cost} measure timing.

\begin{table}[tbp]
\centering
\footnotesize
\setlength{\tabcolsep}{2pt}
\begin{tabular*}{\linewidth}{@{\extracolsep{\fill}}lrrrr@{}}
\toprule
Method & 3 & 13 & 21 & Mean$\pm$std \\
\midrule
\psc{}@8 & 1076 & 1079 & 1078 & 1077.7$\pm$1.5 \\
\psc{}@16 & 1088 & 1090 & 1089 & 1089.0$\pm$1.0 \\
\psc{}@32 & 1096 & 1098 & 1097 & 1097.0$\pm$1.0 \\
\cand{}@16 & 757 & 769 & 765 & 763.7$\pm$6.1 \\
\allaced{}@8 & 1119 & 1122 & 1124 & 1121.7$\pm$2.5 \\
\allaced{}@16 & 1141 & 1146 & 1148 & 1145.0$\pm$3.6 \\
\allaced{}@32 & 1148 & 1151 & 1150 & 1149.7$\pm$1.5 \\
\bottomrule
\end{tabular*}
\caption{Full-GSM8K correct counts for generation seeds 3, 13, and 21. Mean$\pm$std is computed across the three seeds.}
\label{tab:seed}
\end{table}

\subsection{Inference Budgets and Cost}
\label{sec:cost-details}

\begin{table*}[!tbp]
\centering
\footnotesize
\setlength{\tabcolsep}{4pt}
\begin{tabular*}{\textwidth}{@{\extracolsep{\fill}}lrrrrrrr@{}}
\toprule
Method & Correct & Tokens/q & W-gen/q & W-total/q & Gen s/q & Exec s/q & Wall s/q \\
\midrule
\multicolumn{8}{@{}l}{\textit{Direct executable aggregation}} \\
\psc{}@32 & 1098 & 7631.2 & 16837.1 & 53418.1 & 3.93$\pm$0.35 & 0.44$\pm$0.01 & 4.37$\pm$0.34 \\
\psc{}@64 & 1131 & 15931.1 & 38355.5 & 111517.4 & 7.34$\pm$0.71 & 0.88$\pm$0.02 & 8.21$\pm$0.69 \\
\midrule
\multicolumn{8}{@{}l}{\textit{Natural-language and executable evidence}} \\
\allaced{}@8 & 1122 & 6187.9 & 10245.3 & 21241.0 & 2.51$\pm$0.22 & 0.12$\pm$0.01 & 2.63$\pm$0.21 \\
\allaced{}@16 & 1146 & 8979.1 & 19686.2 & 40779.1 & 3.42$\pm$0.31 & 0.22$\pm$0.00 & 3.64$\pm$0.31 \\
\allaced{}@32 & 1151 & 14562.3 & 38574.1 & 79861.5 & 4.7244 & 0.4702 & 5.1947 \\
\bottomrule
\end{tabular*}
\caption{GSM8K inference cost per question. Tokens/q includes input and output. W-gen/q and W-total/q apply model-size weights of 0.5 and 7 to natural-language and code tokens, respectively. W-gen/q counts output tokens only. Timing is mean$\pm$std over seeds 13, 21, and 42, except for the single \allaced{}@32 accounting run. Correct counts are reported evaluation results, not means over timing runs.}
\label{tab:cost}
\end{table*}

\paragraph{Additional SC sampling.}
Doubling the SC budget from $N=16$ to $N=32$ yields 456/1319 correct
answers on GSM8K with 6792.3 exact tokens per question. At comparable
token scale, \cand{}@16 reaches 769 and \allaced{}@8 reaches 1122,
gains of 313 and 666 correct examples. The comparison shows the value
of complementary executable evidence over additional sampling from
the natural-language channel alone.

\paragraph{Executable budgets and cost.}
Table~\ref{tab:cost} separates output-only weighted tokens from the
weighted total and reports generation and execution time individually.
It also adds the $M=32$ operating point to the comparisons in
Table~\ref{tab:main-cost}. Increasing ACED's program budget improves
the correct count from 1122 to 1151, with generation accounting for
most of the recorded time. Exact tokens provide platform-independent
accounting. Timing covers generation and sandbox execution on the
same setup, excluding model loading.

\section{Case Studies}
\label{sec:cases}

Table~\ref{tab:case-audit} illustrates the decisions studied in the
main paper. Panel (a) covers candidate re-selection, executable-only
answer admission, reference retention, and a damaging override.
Panel (b) reports every damaging \allaced{}@16 override on GSM8K.
All examples are drawn from the GSM8K test set.

\begin{table*}[!t]
\centering
\footnotesize
\renewcommand{\arraystretch}{1.08}
\setlength{\tabcolsep}{4pt}
\textbf{(a) Representative outcomes}\\[3pt]
\begin{tabularx}{\textwidth}{@{}L{0.15\textwidth}L{0.51\textwidth}>{\raggedright\arraybackslash}X@{}}
\toprule
Outcome & Problem and decision & Evidence \\
\midrule
Candidate re-selection & Amber, Micah, and Ahito ran 52 miles in total. Amber ran 8 miles. Micah ran 3.5 times what Amber ran. How many miles did Ahito run? Gold 16, SC 14, \cand{} 16. & NL support is $16{:}7$ and $14{:}7$. All 16 programs return 16. \\
Executable-only answer admitted & George has 45\% more pears than bananas. If George has 200 bananas, how many fruits does George have? Gold 490, SC 540, \allaced{} 490. & The gold answer is absent from the NL candidates. All 16 programs return 490. \\
Correct reference retained & Sarah has a rope that is 20 meters long. Her friend wants to buy the rope for \$2 a meter. Sarah plans to use the profit to buy a new rope, which at the store costs \$1.5 a meter. How much money will she have left over after she buys the new rope? Gold 10, SC 10, \allaced{} 10. & The wrong program answer 0 has support 6. The reference answer 10 has program support 5. \\
Damaging override & 90 single-use contacts come in one box and last Pete 45 days. Each box is \$100 and currently 10\% off. If he buys two boxes, how much will each pair of contacts cost? Gold 2, SC 2, \allaced{} 1. & Thirteen programs use a per-contact rather than a per-pair interpretation. \\
\bottomrule
\end{tabularx}
\vspace{8pt}

\textbf{(b) Complete set of damaging \allaced{}@16 overrides}\\[3pt]
\begin{tabularx}{\textwidth}{@{}L{0.16\textwidth}L{0.20\textwidth}r L{0.15\textwidth}>{\raggedright\arraybackslash}X@{}}
\toprule
ID / problem & Gold/SC$\to$ACED & $\Delta$ & Program support & Error mechanism \\
\midrule
27 / canvas-bag profit & $160/160\to288$ & 7.219 & Selected 15, gold 1 & Witnesses treat \$4 as the cost per pack. The gold computation uses \$4 per bag. \\
298 / contact-lens cost per pair & $2/2\to1$ & 10.151 & Selected 13, gold 0 & Witnesses divide the discounted cost by individual contacts rather than pairs. \\
726 / remaining distance & $655/655\to1630$ & 8.848 & Selected 16, gold 0 & Witnesses subtract one day of driving although the problem requires four days. \\
1092 / Newton apple count & $11/11\to10$ & 6.867 & Selected 6, gold 0 & Witnesses omit one retained apple from the sequence of dropped and pocketed apples. \\
1161 / marble price increase & $92/92\to532.466666$ & 6.397 & Selected 6, gold 0 & Witnesses compound a 20\% increase instead of applying each increase to the original price. \\
1290 / halfway return & $2/2\to1$ & 10.080 & Selected 16, gold 0 & Witnesses compute only the direct trip and omit the halfway return detour. \\
\bottomrule
\end{tabularx}
\caption{GSM8K problem-level examples. Panel (a) shows four possible selection outcomes. Panel (b) reports every damaging \allaced{}@16 override and the corresponding program support.}
\label{tab:case-audit}
\end{table*}

\paragraph{Selection, admission, and retention.}
The first example in Table~\ref{tab:case-audit}(a) resolves an SC tie
in favor of a correct sampled answer. Examples 119 and 479 similarly
recover the sampled gold answers 38 and 40 through unanimous
executable support. These decisions improve selection within the
existing answer set.

Executable-only admission expands this set. In example 611, the
natural-language candidates give totals such as 1000, 850, and 800,
while the witnesses return the requested monitor price of 300.
In example 568, all 16 witnesses recover the final percentage 60,
which is absent from the natural-language candidates. Conversely,
examples 256 and 199 retain the correct SC answer despite executable
support for an incorrect alternative.

\paragraph{Semantic target audit.}
All six damaging \allaced{}@16 overrides involve a semantic target
error. Three omit part of the requested scope or event count, two
confuse a unit or denominator, and one misreads a rate.
Table~\ref{tab:case-audit}(b) records each case and its program support.
The programs execute the encoded computation correctly but return a
different quantity from the one requested, connecting the observed
failure mechanism to the motivation for reference-relative selection.

\section{Reproducibility Details}
\label{sec:reproducibility}

The reproduction details follow the inference pipeline from generation
and prompts to baseline selection, answer normalization, and execution.

\subsection{Generation and Execution Settings}

Table~\ref{tab:generation-settings} summarizes the models, sample budgets,
and execution settings.

\begin{table}[!htbp]
\centering
\footnotesize
\setlength{\tabcolsep}{3pt}
\begin{tabularx}{\linewidth}{@{}L{0.30\linewidth}>{\raggedright\arraybackslash}X@{}}
\toprule
Setting & Value \\
\midrule
NL candidate model & Qwen/Qwen2.5-0.5B-Instruct \\
Stronger SC reference model & Qwen/Qwen2.5-7B-Instruct \\
Code witness / direct \psc{} model & Qwen/Qwen2.5-Coder-7B-Instruct \\
Reported sample counts & $N=16$ and $M\in\{8,16,32\}$ \\
Generation limits & 256 maximum new tokens for natural-language candidates and 192 for \method{} witnesses and direct \psc{} programs \\
Code prompt limit & 1,536 tokens for code-generation prompts \\
Recorded timing & CUDA with bf16 and batch size 8 on full GSM8K over seeds 13, 21, and 42. Timing includes generation and sandbox execution and excludes model loading. \\
Timing scope & Every timing comparison uses the same local setup; exact token counts provide the hardware-independent cost record. \\
Program execution & callable \texttt{solve()}, arithmetic imports, restricted builtins, 2.0 second timeout \\
\bottomrule
\end{tabularx}
\caption{Models, generation limits, and execution settings. Baseline-specific settings appear in Appendix~\ref{sec:baseline-notes}.}
\label{tab:generation-settings}
\end{table}

\begin{table}[!htbp]
\centering
\footnotesize
\setlength{\tabcolsep}{3pt}
\begin{tabularx}{\linewidth}{@{}L{0.27\linewidth}>{\raggedright\arraybackslash}X@{}}
\toprule
Component & Implementation entry point \\
\midrule
NL prompt and extraction & \texttt{src/phase\_a\_tango.py}: \texttt{render\_prompt}, \texttt{extract\_gsm8k\_answer}, \texttt{normalize\_numeric\_answer} \\
Direct \psc{} prompt & \url{src/exact_acv/exact_acv.py}: \texttt{build\_solver\_prompt}, and \url{src/race_acv/prompt_program_baseline.py}: \texttt{build\_program\_prompt} \\
\method{} witness prompt & \url{src/race_acv/prompt_program_baseline.py}: \texttt{build\_program\_prompt}, style \texttt{repair\_problem} \\
\prove{} prompt & \url{src/race_acv/prove_baseline.py}: \texttt{build\_prove\_prompt} \\
\acr{} prompt & \url{src/race_acv/acr_option_baseline.py}: \texttt{build\_prompt} \\
Program extraction, filtering, execution & \url{src/exact_acv/exact_acv.py}: \texttt{extract\_code}, \texttt{is\_suspicious\_program}, \texttt{execute\_code} \\
\bottomrule
\end{tabularx}
\caption{Source-level implementation hooks used for the reported experiments. Paths are relative to the implementation root.}
\label{tab:implementation-hooks}
\end{table}

\subsection{Prompt Interfaces}
\label{sec:prompts}

\method{} witnesses are \emph{problem-only}. They receive the problem
text but not candidate answers, the SC reference, gold labels, or
natural-language rationales. Direct \psc{} is also problem-only but
uses a separate direct-solver prompt. Both program streams use the same
code extraction, safety checks, execution sandbox, and answer
normalization policy. The distinct prompts produce separate program
pools. The controls in Tables~\ref{tab:ablation}
and~\ref{tab:m-scaling-ablation} apply both selection rules to the
same ACED pool. In contrast, \prove{} translates a sampled
candidate solution into a program, while \acr{} asks a teacher model
to select among answer options.

Table~\ref{tab:prompt-problem-only} presents the prompt text in two
groups. Panel (a) receives only the problem, whereas panel (b)
additionally receives a sampled solution or answer options.
The template wording is preserved verbatim.

\begin{table*}[!tbp]
\centering
\footnotesize
\renewcommand{\arraystretch}{1.12}
\setlength{\tabcolsep}{4pt}
\begin{tabularx}{\textwidth}{@{}L{0.16\textwidth}>{\raggedright\arraybackslash}X@{}}
\toprule
Interface & Template text \\
\midrule
\multicolumn{2}{@{}l}{\textit{(a) Problem-only interfaces}} \\
\addlinespace[3pt]
NL candidate generator & Solve the following math problem. Reason step by step and put the final answer at the end. Problem: \{question\}. \\
\psc{} & You are given a grade-school math word problem. Write Python code that defines a function \texttt{solve()}. Rules: compute the answer from the quantities in the problem; do not hard-code the final answer; use only arithmetic and optional math, fractions, or decimal; return a single number from \texttt{solve()}; output only Python code. Problem: \{question\}. Python code: \\
\method{} witness & Solve this math word problem independently. Write only Python code. The code must define \texttt{solve()} and return one numeric answer. Before coding, audit the wording of the final question and identify the exact requested quantity. Then define variables and equations from the problem and compute that requested quantity. Do not hard-code the final answer. Do not return a subtotal, rate, leftover, or intermediate value unless it is exactly what is asked. Problem: \{question\}. Python code: \\
\midrule
\multicolumn{2}{@{}l}{\textit{(b) Candidate-conditioned interfaces}} \\
\addlinespace[3pt]
\prove{} & You are verifying a grade-school math solution by translating its arithmetic into Python. Write only Python code. The code must define \texttt{solve()} and return one numeric answer. Follow the quantitative steps in the provided solution. Do not simply return a final literal answer; derive it from the quantities and operations. Problem: \{question\}. Candidate solution: \{candidate\_solution\}. Python code: \\
\acr{} & You are selecting the most plausible final answer for a grade-school math problem. Choose exactly one option. Reply with only the option letter. Problem: \{question\}. Candidate final answers: A. \{answer\_1\} (sample count: \{count\_1\}/16), B. \{answer\_2\} (sample count: \{count\_2\}/16), \ldots. Best option: \\
\bottomrule
\end{tabularx}
\caption{Prompt templates used by the five interfaces.
Braced placeholders indicate inserted fields.
The ellipsis in the \acr{} template denotes further answer options
in the same format.}
\label{tab:prompt-problem-only}
\label{tab:prompt-conditioned}
\end{table*}

\subsection{Baseline Reproduction Details}
\label{sec:baseline-notes}

The baselines share the final-answer normalization and reporting
interface. Their prompts, extraction procedures, decision rules, and
fallbacks are specified below.

\paragraph{\acr{}.}
Our \acr{} baseline adapts the published method by combining
natural-language sample counts with five
teacher-model votes over the top eight answer clusters.
Our implementation queries every problem and weights the option-vote
support by 0.25. The published method instead selectively queries
uncertain cases and sums the two support terms
without this relative weighting.
Table~\ref{tab:acr-provenance} records the scoring rule and fallback.
The five votes per question give 6595 calls on full GSM8K, of which
99.5\% produce a parsed option vote.

\begin{table}[!htbp]
\centering
\footnotesize
\setlength{\tabcolsep}{3pt}
\begin{tabular}{@{}>{\raggedright\arraybackslash}p{0.34\linewidth}>{\raggedright\arraybackslash}p{0.58\linewidth}@{}}
\toprule
Item & Setting \\
\midrule
Baseline type & Answer-option rescoring baseline \\
Publication & DOI: 10.3390/app15179587 \\
Candidate domain & Candidate-only, stored $N=16$ natural-language answer clusters \\
Teacher model & Qwen/Qwen2.5-Coder-7B-Instruct code-model teacher \\
Options / votes & Top $K=8$ answer options and 5 teacher votes per question \\
Weight and rule & Weight 0.25. Select by $n_{\mathrm{NL}}/16 + 0.25\cdot v_{\mathrm{teacher}}/5$. \\
Fallback & If no teacher vote parses, use the natural-language-count / \scbase{} behavior \\
Parsed votes & 6562/6595 option votes and 1316/1319 questions with at least one parsed vote \\
Result & 410/1319 correct, 81 damaging overrides, and 72 fixes \\
\bottomrule
\end{tabular}
\caption{\acr{} configuration under the paper's full-GSM8K candidate reporting interface.}
\label{tab:acr-provenance}
\end{table}

The 72 fixes and 81 damaging overrides account for the nine-example
decrease from SC through Eq.~\eqref{eq:fix-damage-identity}.

\paragraph{\prove{}.}
\prove{} uses the candidate-conditioned verification pipeline summarized in Table~\ref{tab:prove-provenance}. For each problem, it generates 16 CoTs with the 0.5B natural-language model and translates every candidate solution into a Python verifier with the 7B code model. If no candidate verifies, it returns the extracted CoT majority. On full GSM8K, this baseline executes 21,104 translated programs. Of these, 19,321 are valid (91.6\%), 762 are flagged suspicious (3.6\%), and 4,963 candidate traces verify (23.5\%). The method uses its fallback on 368/1319 examples. The resulting row is 654/1319 with 9 damaging overrides and 244 fixes.

\begin{table}[!htbp]
\centering
\footnotesize
\setlength{\tabcolsep}{3pt}
\begin{tabular}{@{}>{\raggedright\arraybackslash}p{0.38\linewidth}>{\raggedright\arraybackslash}p{0.54\linewidth}@{}}
\toprule
Argument & Value \\
\midrule
Repository & \url{https://github.com/declare-lab/prove} \\
Commit & \begin{tabular}[t]{@{}l@{}}a5a8f3feabf05b340df4d9cf\\d72b2bf64bec6345\end{tabular} \\
Entry / pipeline & \texttt{eval.py}, pipeline \texttt{prove} \\
Dataset / samples & \texttt{gsm8k}, \texttt{num\_cot=16} \\
CoT generation & \texttt{cot} template, \texttt{qwen2\_05b}, temperature 0.7, max tokens 1024 \\
Answer extraction & \texttt{extract} template, \texttt{qwen25coder\_7b}, temperature 0.0, max tokens 32 \\
Program translation & \texttt{output2pot}, \texttt{qwen25coder\_7b}, temperature 0.0, max tokens 1024 \\
\bottomrule
\end{tabular}
\caption{\prove{} repository configuration used for the candidate-domain baseline.}
\label{tab:prove-provenance}
\end{table}

\paragraph{\psc{} and additional SC sampling.}
\psc{} uses the same code model and execution interface as ACED,
generates a separate pool with its direct-solver prompt, and returns
the majority program answer. The fixed-pool comparison is defined in
Appendix~\ref{sec:source-selector}. Additional \scbase{}@32 sampling
uses the \scbase{}@16 generator with twice the sample budget.

\subsection{Normalization and Filtering}
\label{sec:normalization-filtering}

\paragraph{Execution and normalization counts.}
Table~\ref{tab:normalization-audit} partitions the generated outputs.
Safety and validation checks determine which rows enter
$M_{\mathrm{valid}}$. Normalization then assigns each retained row
to a numeric answer count $v(a)$ or the non-selectable null bucket
$\varnothing$.

\begin{table}[!htbp]
\centering
\footnotesize
\setlength{\tabcolsep}{2pt}
\begin{tabularx}{\linewidth}{@{}>{\raggedright\arraybackslash}Xrrr@{}}
\toprule
Output category & NL & ACED & \psc{} \\
\midrule
Outputs total & 21,104 & 42,208 & 42,208 \\
Excluded before $M_{\mathrm{valid}}$ & 0 & 5,394 & 2,478 \\
Retained rows ($N$ or $M_{\mathrm{valid}}$) & 21,104 & 36,814 & 39,730 \\
Numeric answer rows & 21,104 & 34,461 & 39,673 \\
Fallback non-numeric keys & 0 & 0 & 0 \\
No-answer null-bucket rows & 0 & 2,353 & 57 \\
Fallback/no-answer gold matches & 0 & 0 & 0 \\
Normalization collision keys & 156 & 179 & 242 \\
Rows involved in collision keys & 14,619 & 28,480 & 33,294 \\
Correct rows involved in collision keys & 3,137 & 26,019 & 28,481 \\
\bottomrule
\end{tabularx}
\caption{Full-GSM8K output and normalization counts. ACED and \psc{} use $M=32$. Collision keys receive multiple raw forms that normalize to the same numeric answer.}
\label{tab:normalization-audit}
\end{table}

\paragraph{Answer normalization.}
All answer sources use the same final-answer normalization before voting or scoring. In the ACED natural-language branch, answers are extracted by first looking for a GSM8K-style final marker and otherwise using the last numeric value. ProVe uses the extraction interface in Table~\ref{tab:prove-provenance}, while ACR parses answer-option votes. Numeric strings have commas and surrounding whitespace removed and are parsed as floats.

\paragraph{Numeric canonicalization.}
Values within $10^{-6}$ of an integer are canonicalized to an integer
string. Other values are rounded to at most six decimals with trailing
zeros stripped. Program outputs additionally recognize
\texttt{Fraction(a,b)} representations and otherwise use the last
numeric value in the returned object or string. A nonnumeric residual
may be retained for audit bookkeeping, but it is excluded from the
selectable answer domain and enters the null bucket.
The collision counts in Table~\ref{tab:normalization-audit} record
canonical answers that receive multiple equivalent surface forms, such
as \texttt{5}, \texttt{5.0}, and \texttt{5.00}. These counts measure
equivalent-form aggregation. All selectable keys in the reported
full-GSM8K runs are numeric.

\paragraph{Code extraction.}
When a fenced Python block is present, the parser retains that block. It removes a leading \texttt{Python code:} marker, begins at the first line starting with \texttt{import}, \texttt{from}, or \texttt{def solve}, and truncates common explanation tails such as \texttt{Explanation:} and \texttt{Output:}. These rules remove surrounding model commentary without changing the executable body.

\paragraph{Program filtering.}
Filtering is applied before execution and does not use candidate answers or labels. A program is rejected if it contains file, process, network, operating-system, or double-underscore access. The explicit blocked patterns include \texttt{open(}, \texttt{input(}, \texttt{exec(}, \texttt{eval(}, \texttt{compile(}, \texttt{subprocess}, \texttt{socket}, \texttt{os.}, and \texttt{sys.}. The filter also rejects a direct numeric-literal return. When the problem contains numeric quantities, the code must use at least $\min(2,k)$ of the $k$ quantities found in the problem text. This last rule removes programs that ignore most of the available numerical information.

\begin{table}[!htbp]
\centering
\footnotesize
\setlength{\tabcolsep}{3pt}
\begin{tabular}{@{}L{0.28\linewidth}L{0.64\linewidth}@{}}
\toprule
Sandbox component & Allowed interface \\
\midrule
Imports & \texttt{math}, \texttt{fractions}, and \texttt{decimal} \\
Builtins & \texttt{abs}, \texttt{min}, \texttt{max}, \texttt{sum}, \texttt{round}, \texttt{range}, \texttt{len}, \texttt{float}, \texttt{int}, \texttt{str}, \texttt{pow}, and \texttt{print} \\
Primary entry point & Callable \texttt{solve()} returning the answer \\
Fallback result names & \texttt{result}, \texttt{answer}, \texttt{ans}, \texttt{total}, \texttt{total\_cost}, \texttt{final}, and \texttt{final\_answer} \\
Execution limit & 2.0 seconds per program \\
\bottomrule
\end{tabular}
\caption{Restricted execution interface for generated Python programs.}
\label{tab:execution-sandbox}
\end{table}

Programs run in an isolated Python process under the interface in Table~\ref{tab:execution-sandbox}. A callable \texttt{solve()} is the expected entry point. The fallback result names are checked only when no callable entry point is available. Programs that time out or fail validation are excluded before $M_{\mathrm{valid}}$ is formed, while retained executions without a normalized numeric answer enter the non-selectable null bucket defined in the evidence-pooling derivation.

\FloatBarrier

\section{Experimental Environment and Computational Budget}
\label{sec:environment-budget}

\subsection{Hardware and Model Configuration}
\label{sec:hardware-configuration}

The GPU infrastructure comprised two NVIDIA A100 GPUs with 80\,GB
of memory each. Each recorded timing job used one GPU, with bf16
precision. Table~\ref{tab:hardware-configuration} summarizes the
hardware, nominal model sizes, adapter settings, and sample budgets.

\begin{table}[!htbp]
\centering
\footnotesize
\setlength{\tabcolsep}{3pt}
\begin{tabularx}{\linewidth}{@{}L{0.39\linewidth}>{\raggedright\arraybackslash}X@{}}
\toprule
Item & Configuration \\
\midrule
GPU infrastructure & $2\times$ NVIDIA A100 \\
GPU memory & 80\,GB per GPU \\
GPUs per timing job & 1 \\
Numerical precision & bf16 \\
Main NL model & Qwen2.5-0.5B-Instruct (nominally 0.5B parameters) \\
Stronger NL model & Qwen2.5-7B-Instruct (nominally 7B parameters) \\
Code model & Qwen2.5-Coder-7B-Instruct (nominally 7B parameters) \\
Main-reference adapter & CE LoRA, rank 8, $\alpha=16$, \mbox{dropout 0} \\
Main NL budget & $N=16$ candidates per question \\
ACED code budget & $M\in\{8,16,32\}$ witnesses per question \\
Direct \psc{} budget & $M\in\{8,16,32,64\}$ programs per question \\
Peak allocated GPU memory & Not recorded \\
\bottomrule
\end{tabularx}
\caption{Hardware and model configuration. GPU memory denotes device capacity, not measured peak allocation.}
\label{tab:hardware-configuration}
\end{table}

\FloatBarrier
\subsection{Software and Execution Configuration}
\label{sec:software-configuration}

Inference was implemented with PyTorch and Hugging Face Transformers.
Table~\ref{tab:reproduction-environment} specifies the accompanying
reproduction environment. The complete original Python, CUDA toolkit,
NVIDIA driver, and library-version snapshot was not retained, so these
versions describe the reproduction environment rather than a verified
historical snapshot. The saved adapter metadata records \mbox{PEFT 0.19.1}.

\begin{table}[!htbp]
\centering
\footnotesize
\setlength{\tabcolsep}{3pt}
\begin{tabularx}{\linewidth}{@{}Xr@{}}
\toprule
Package & Version \\
\midrule
PyTorch & 2.7.1 \\
Transformers & 4.57.1 \\
Accelerate & 1.13.0 \\
Datasets & 4.8.5 \\
NumPy & 2.2.6 \\
Pandas & 2.3.3 \\
SciPy & 1.16.3 \\
PyYAML & 6.0.3 \\
\bottomrule
\end{tabularx}
\caption{Package versions specified for reproduction.}
\label{tab:reproduction-environment}
\end{table}

\FloatBarrier
\subsection{Computational Accounting}
\label{sec:computational-accounting}

Appendix~\ref{sec:cost-details} reports input-plus-output token counts,
model-size-weighted token counts, and generation-plus-execution time
per question. The recorded timing jobs use one GPU and exclude model
loading. These measurements characterize the reported inference
configurations rather than a total GPU-hour budget across all
experiments.

\section{Ethical Considerations}
\label{sec:ethical-considerations}

\method{} supports research on reliable mathematical answer selection.
Its correction and damage analysis makes both beneficial and harmful
answer changes explicit. In applications, executable agreement should
be treated as evidence rather than a correctness guarantee, since
overreliance on an incorrect answer can propagate errors into later
decisions. Generated code also warrants controlled execution. The
restricted interfaces, filtering, and timeouts in
Appendix~\ref{sec:normalization-filtering} provide execution precautions,
and sensitive applications should retain independent answer checks.
Repeated inference consumes computational resources. Reporting token
budgets and runtime supports informed resource allocation, while
latency reductions alone do not establish energy or emissions savings.

\section{AI Assistance}
\label{sec:ai-assistance}

AI tools assisted with manuscript drafting and organization, literature search and code implementation. The authors made the final research decisions and retain full responsibility for the accuracy and integrity of the methods, results, code, figures, and references.

\end{document}